\documentclass{article}

\usepackage{epsfig}
\usepackage{graphicx}
\usepackage{amsmath}
\usepackage{algorithm}
\usepackage{algorithmic}
\usepackage{amssymb}
\usepackage{amsthm}
\usepackage{cmap}
\usepackage{appendix}
\usepackage[utf8]{inputenc} 
\usepackage[T1]{fontenc}    
\usepackage{hyperref}       
\usepackage{url}            
\usepackage{booktabs}       
\usepackage{amsfonts}       
\usepackage{nicefrac}       
\usepackage{microtype}      

\usepackage[nonatbib,final]{nips_2016} 

\newcommand{\comment}[1]{}
\newcommand{\floor}[1]{\lfloor #1 \rfloor}
\newcommand{\argmin}{\operatornamewithlimits{argmin}}

\newcommand{\mysection}[1]{\vspace{0mm}\section{#1}\vspace{0mm}}
\newcommand{\mysubsection}[1]{\vspace{0mm}\subsection{#1}\vspace{0mm}}

\newcommand{\myparagraph}[1]{\vspace{0mm}\paragraph{#1}}
\newcommand{\mycaption}[1]{\vspace{0mm}\caption{#1}\vspace{0mm}}
\newcommand{\imagePath}{./images}

\newtheorem{proposition}{Proposition} 
 
\newtheorem{lemma}{Lemma}
\newtheorem{property}{Property}

\title{Truncated Max-of-Convex Models}

\author{
  Pankaj Pansari\\
  University of Oxford\\
  \texttt{pankaj@robots.ox.ac.uk}\\
  \And
  M. Pawan Kumar\\
  University of Oxford\\
  \texttt{pawan@robots.ox.ac.uk}
}

\begin{document}

\maketitle

\begin{abstract}
Truncated convex models (TCM) are a special case of pairwise random fields that have been widely used in computer vision. However, by restricting the order of the potentials to be at most two, they fail to capture useful image statistics. We propose a natural generalization of TCM to high-order random fields, which we call {\em truncated max-of-convex models} (TMCM). The energy function of TMCM consists of two types of potentials: (i) unary potential, which has no restriction on its form; and (ii) clique potential, which is the sum of the $m$ largest truncated convex distances over all label pairs in a clique. The use of a convex distance function encourages smoothness, while truncation allows for discontinuities in the labeling. By using $m > 1$, TMCM provides robustness towards errors in the definition of the cliques.  In order to minimize the energy function of a TMCM over all possible labelings, we design an efficient $st$-{\sc mincut} based range expansion algorithm. We prove the accuracy of our algorithm by establishing strong multiplicative bounds for several special cases of interest. Using standard real data sets, we demonstrate the benefit of our high-order TMCM over pairwise TCM, as well as the benefit of our range expansion algorithm over other $st$-{\sc mincut} based approaches.  
\end{abstract}

\mysection{Introduction}
Truncated convex models (TCM) are a special case of pairwise random fields that have been widely used for low-level vision applications. A TCM is defined over a set of random variables, each of which can be assigned a value from a finite, discrete and ordered label set.  In addition, a TCM also specifies a neighborhood relationship over the random variables. An assignment of values to all the variables is referred to as a labeling. In order to quantitatively distinguish the labelings, a TCM specifies an energy function that consists of unary and pairwise potentials.

\begin{figure}
	\centering
\begin{tabular}{c}
	\includegraphics[scale = 0.3]{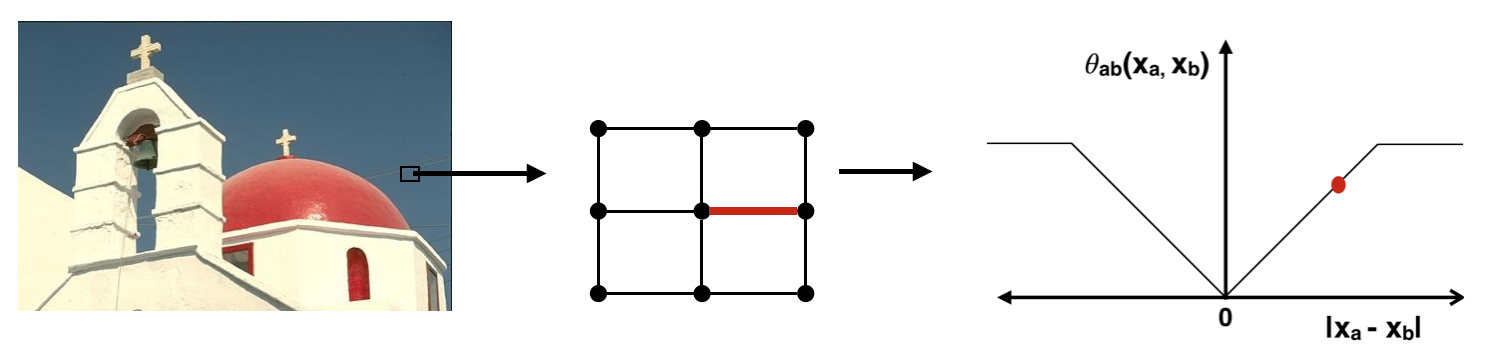} \\
	(a) Truncated convex model \\
	\includegraphics[scale = 0.3]{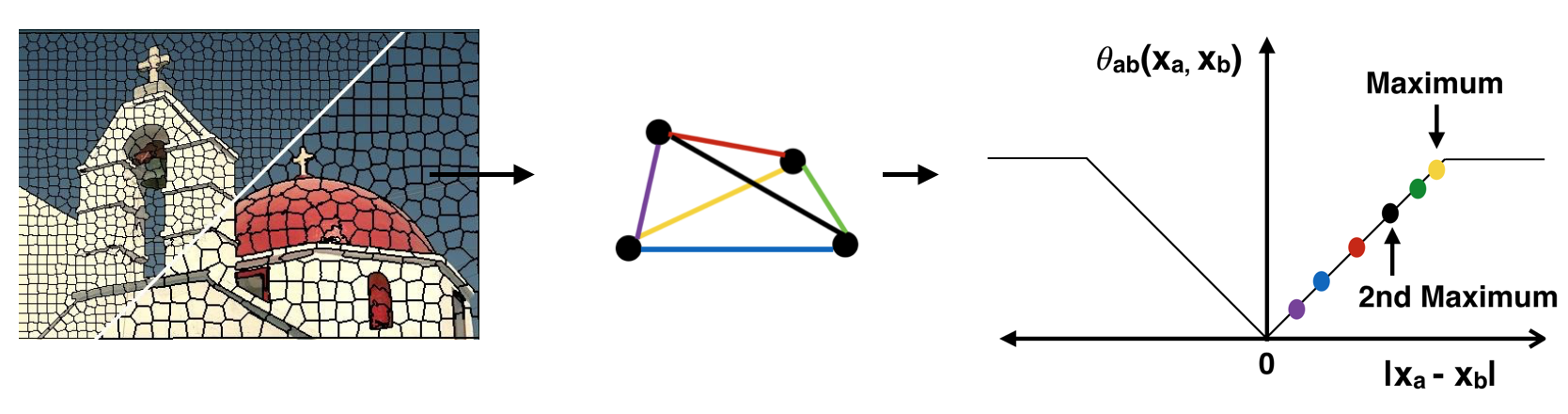} \\ 
	(b) Truncated max-of-convex model \\
\end{tabular}
\vspace{8mm}
\mycaption{\footnotesize \em TMCM as generalization of TCM. In (a), given an image, TCM considers pairwise 4-neighborhood relationships and uses truncated convex distance function for pairwise potential. In (b), TMCM considers superpixels as cliques. The clique potential for $m$ =2 is the sum of the first and second maximum over all the pairwise truncated convex distances such that no variable is used more than once.}  
\vspace{-5mm}
\label{fig:tmcm_concept}
\end{figure}

Given an input, the output is obtained by minimizing the energy function of a TCM over all possible labelings. While this is an NP-hard problem, several approximate algorithms have been proposed in the literature~\cite{boykovpami01,chekurisoda01,guptastoc00,kolmogorovpami06,komodakisiccv07,komodakiscvpr07,kumar2011improved, sontag2012tightening, vekslercvpr07}, which provide accurate solutions in practice~\cite{szeliskipami08}. 

Since we cannot reasonably expect to improve the optimization of TCM, any failure cases must be addressed by modifying the model itself to better capture image statistics. To this end, we propose to address one of the main deficiencies of TCM, namely, the restriction to potentials of order at most two. Specifically, we propose a natural generalization of TCM to high-order random fields, which we refer to as {\em truncated max-of-convex models} (TMCM).  Similar to TCM, our model places no restrictions on the unary potentials.  Furthermore, unlike TCM, it allows us to define clique potentials over an arbitrary number of random variables.  The value of the clique potential is proportional to the sum of the $m$ largest truncated convex distances computed over disjoint pairs of random variables in the clique. Here, disjoint pairs imply that the label of no random variable is used more than once to compute the value of the clique potential. Figure~\ref{fig:tmcm_concept} demonstrates how TMCM differs from TCM. The exact mathematical form of the TMCM energy function will be presented in section ~\ref{sec:TMCM}.  The term $m$ is a positive integer that is less than or equal to half the number of variables in the smallest clique.  Importantly, the constant of proportionality for each clique potential can depend on the input corresponding to all the random variables in the clique. This can help capture more interesting image statistics, which in turn can lead to a more desirable output. For example, in image denoising, the clique weights can depend on the variance of intensity over a superpixel (group of pixels with similar semantic and perceptual characteristics).  

In order to enable the use of TMCMs in practice, we require an efficient and accurate energy minimization algorithm that can compute the output for a given input. To this end, we extend the range expansion algorithm for TCM to deal with arbitrary sized clique potentials. Our algorithm retains the desirable property of iteratively solving an $st$-{\sc mincut} problem over an appropriate directed graph (where the number of vertices and arcs grows linearly with the number of random variables and cliques, and quadratically with the number of labels). As the $st$-{\sc mincut} problem lends itself to several fast algorithms~\cite{boykovpami04}, this makes our overall approach computationally efficient. Furthermore, we provide strong theoretical guarantees on the quality of the solution for several special cases of interest, which establishes its accuracy. Our multiplicative bounds are better than the baselines for cases where comparison is possible. Using standard real data sets, we show the benefit of high-order TMCM over pairwise TCM, as well as the advantage of our range expansion algorithm over other $st$-{\sc mincut} based approaches.

\mysection{Related Work}
Pairwise TCM offer a natural framework to capture low-level cues for vision problems such as image denoising, stereo correspondence, segmentation and optical flow~\cite{szeliskipami08}. However, the restriction to pairwise potentials limits their representational power.  

For the past few years, there has been a growing interest in high-order models. Though other inference algorithms such as message passing are possible~\cite{felzenszwalb2006efficient, kolmogorov2006convergent, komodakis2007mrf, meshi2010learning, tarlow2010hop, wainwright2005map}, in this work our focus is on models that admit efficient $st$-{\sc mincut} based solutions and provide strong theoretical guarantees on the quality of the solution. One early work was the $P^n$ Potts model~\cite{kohlicvpr07}, which encourages label consistency over a set of random variables. This work was extended in~\cite{kohlicvpr08}, which introduced robustness in the $P^n$ Potts model by taking into account the number of random variables that disagreed with the majority label of a clique. Both the $P^n$ Potts model and its robust version lend themselves to efficient optimization via the expansion algorithm~\cite{boykovpami01}, which solves one $st$-{\sc mincut} problem at each iteration. The expansion algorithm provides multiplicative bounds~\cite{gouldcvpr09}, which measure the quality of the estimated labeling with respect to the optimal one. Our work generalizes both the models, as well as the corresponding expansion algorithm. Specifically, when the truncation factor of our models is set to 1, we recover the robust $P^n$ model.  Furthermore, a suitable setting of the range expansion algorithm (setting the interval length to 1) recovers the expansion algorithm. 

Jegelka and Bilmes \cite{jegelka2011submodularity} introduced a nonsubmodular high-order model which is based on edge cooperation and is optimizable using $st$-{\sc mincut}, but the algorithm has weak approximation bounds. Delong et al.\ \cite{delongijcv12,delongcvpr10} proposed a clique potential based on label costs that can also be handled via the expansion algorithm.  However, unlike the robust $P^n$ Potts model, their model provides additive bounds that are not invariant to reparameterizations of the energy function. This theoretical limitation is addressed by the recent work of Dokania and Kumar~\cite{dokaniaiccv15} on parsimonious labeling. Here, the clique potentials are defined as being proportional to a diversity function of the set of unique labels present in the clique labeling. Our work can be thought of as being complementary to parsimonious labeling. Specifically, while parsimonious labeling is an extension of pairwise metric labeling to high-order models, our work is an extension of TCM. The only metric that also results in a TCM is the truncated linear distance. As our experiments will demonstrate, our specialized range expansion algorithm provides significantly better results for truncated max-of-linear models compared to the hierarchical $st$-{\sc mincut} approach of~\cite{dokaniaiccv15}.  

We note that there have been several works that deal with more general high-order potentials and design $st$-{\sc mincut} style solutions for them. For example, Fix et al.\ \cite{fixcvpr14} use the submodular max-flow algorithm~\cite{kolmogorovdam12}, while Arora et al.\ \cite{aroracvpr14} use generic cuts. However, the resulting algorithms are exponential in the size of the cliques, which prevents their use in applications that require very high-order cliques (with hundreds or even thousands of random variables). A notable exception to this is the work of Ladicky et al.\ \cite{ladickyeccv10}, who proposed a co-occurrence based clique potential whose only requirement is that it should increase monotonically with the set of unique labels present in the clique labeling. However, the use of such a general clique potential still results in an inaccurate energy minimization algorithm, as will be seen in our experimental comparison.  

\mysection{Truncated Convex Models}

A TCM is a random field defined by a set of discrete random variables
${\bf X} = \{X_a, a \in {\cal V}\}$, and a neighborhood relationship ${\cal E}$ over them (that is, $X_a$ and $X_b$ are neighboring random
variables if $(a,b) \in {\cal E}$).
Each random variable can take a value from a finite label set ${\bf L}$, which
is assumed to be ordered to enable the use of convex distance functions. Without loss of generality, we define ${\cal V} = \{1,2,\cdots,n\}$ and ${\bf L} = \{1,2,\cdots,h\}$.

An assignment of values to all the random variables ${\bf x} \in {\bf L}^n$ is referred to as a labeling. To quantitatively distinguish the $h^n$ possible labelings, a TCM defines an energy function that consists of two types of potentials. First, the unary potentials $\theta_a(x_a)$ that depends on the label $x_a$ of one random variable $X_a$. Second, the pairwise potentials $\theta_{ab}(x_a,x_b)$ that depends on the labels $x_a$ and $x_b$ of a pair of neighboring random variables $(X_a,X_b)$. There are no restrictions on the form of the unary potentials. However, the pairwise potentials are defined using a truncated convex distance function over the label set.

To provide a formal specification of the pairwise potentials, we require some definitions. We denote a convex distance function by $d: \mathbb{Z} \rightarrow \mathbb{R}$ (where $\mathbb{Z}$ is the set of integers and $\mathbb{R}$ is the set of real numbers).  Recall that a convex distance function satisfies the following properties: (i) $d(y) \geq 0$ for all $y \in \mathbb{Z}$ and $d(0) = 0$; (ii) $d(y) = d(-y)$ for all $y \in \mathbb{Z}$; and (iii) $d(y+1)-2d(y)+d(y-1) \geq 0$ for all $y \in \mathbb{Z}$. Note that the above properties also imply that $d(y) \geq d(z)$ if $|y| \geq |z|$, for all $y,z \in \mathbb{Z}$.  Examples of convex distance functions include the linear (that is, $d(y) = |y|$) and the quadratic distance function (that is, $d(y) = y^2$) distance.

Given two labels $l_i,l_j \in {\bf L}$, we use a convex function $d(\cdot)$ to measure the distance between them as $d(l_i-l_j)$, thereby encouraging smooth labelings. In order to prevent the overpenalization of the discontinuities in an image, it is common practice to truncate the convex distance function~\cite{boykovpami01,kumarnips08,vekslercvpr07}. Formally, a truncated convex function is defined as $\min\{d(\cdot),M\}$, where $M$ is the truncation factor. We now define the pairwise potential as $\theta_{ab}(x_a,x_b) = \omega_{ab}\min\{d(x_a-x_b),M\}$, where $\omega_{ab}$ is a (data-dependent) non-negative constant of proportionality.

Hence, a TCM specifies an energy function $E(\cdot)$ over the labelings ${\bf x} \in {\bf L}^n$ as follows: 

\begin{equation} 
\small E({\bf x}) = \sum_{a \in {\cal V}} \theta_a(x_a) + \sum_{(a,b) \in {\cal E}} \omega_{ab}\min\{d(x_a-x_b),M\}.  
\end{equation} 
The unary potentials are arbitrary, the edge weights $\omega_{ab}$ are non-negative, $d(\cdot)$ is a convex function and $M \geq 0$ is the truncation factor.  Given an input (which provides the values of the unary potentials and the edge weights), the desired output is obtained by solving the following optimization problem: $\min_{{\bf x} \in {\bf L}^n} E({\bf x})$. While this optimization problem is NP-hard, we can obtain an accurate approximate solution by using the efficient range expansion algorithm~\cite{kumarnips08}, as well as several other approaches based on $st$-{\sc mincut}~\cite{boykovpami01,guptastoc00,komodakiscvpr07,vekslercvpr07} and linear programming~\cite{chekurisoda01,kolmogorovpami06,komodakisiccv07}.  

\mysection{Truncated Max-of-Convex Models}
\label{sec:TMCM}

We now present a natural generalization of TCM to high-order random fields, which define potentials over random variables that form a clique (where all the random variables in a clique are neighbors of each other). Importantly, we do not place any restriction on the size of the clique. 

\myparagraph{\bf Truncated Max-of-Convex Potentials.}  Consider a high-order clique consisting of the random variables ${\bf X}_{\bf c} = \{X_a, a \in {\bf c} \subseteq {\cal V}\}$.
We denote a labeling of the clique as ${\bf x}_{\bf c} \in {\bf L}^{c}$, where we have used the shorthand $c = |{\bf c}|$ to denote the size of the clique.
In order to specify the value of the clique potential for the labeling ${\bf x}_{\bf c}$ we
require a sorted list of the (not necessarily unique) labels present in ${\bf x}_{\bf c}$. We denote this sorted list by ${\bf p}({\bf x}_{\bf c})$
and access its $i$-th element as $p_i({\bf x}_{\bf c})$. For example, consider a clique consisting of random variables ${\bf X}_{\bf c} = \{X_1,X_2,X_3,X_4,X_5,X_6\}$.
If the number of labels is $h = 10$, then one of the putative labelings of the clique is ${\bf x}_{\bf c} = \{3,2,1,4,1,3\}$ (that is, $X_1$ takes the value $3$,
$X_2$ takes the value $2$ and so on). For this labeling, ${\bf p}({\bf x}_{\bf c}) = \{1,1,2,3,3,4\}$. The value of $p_1({\bf x}_{\bf c})$ and
$p_2({\bf x}_{\bf c})$ is $1$, the value of $p_3({\bf x}_{\bf c})$ is $2$ and so on. Given a convex function $d(\cdot)$, a truncation factor $M$ and an integer
$m \in [0,\floor{c/2}]$, the clique potential $\theta_{\bf c}(\cdot)$ is defined as

\begin{equation}
\small \theta_{\bf c}({\bf x}_{\bf c}) = \omega_{\bf c} \sum_{i=1}^m \min\{d(p_i({\bf x}_{\bf c})-p_{c-i+1}({\bf x}_{\bf c})),M\}.
\label{eq:maxPotential}
\end{equation}
Here, $\omega_{\bf c} \geq 0$ is the clique weight that does not depend on the labeling. However, it can be chosen based on the observed data - for instance, we may want to assign small weights to cliques with large variance of intensity or disparity. The term inside the summation is the truncated
value of the $i$-th largest distance between the labels of all pairs of random variables within the clique, subject to the constraint that the label of no random variable is used more than once in the computation of the clique potential value. In other words, our clique potential is proportional to the
sum of the truncation of the $m$ largest convex distance functions over disjoint pairs of random variables. 

Given an input, the desired output is obtained by solving the following optimization problem: $\min_{{\bf x} \in {\bf L}^n} E({\bf x})$. Note that TMCM is a generalization of the $P^n$ Potts model~\cite{kohlicvpr07} ($m$ = 1, $M$ = 1) as well as its robust version~\cite{kohlicvpr08} ($m > $  1, $M$ = 1). Furthermore, it is complementary to the recently proposed parsimonious labeling, which generalizes metric labeling.  Henceforth, we assume the unary potentials are non-negative. This assumption is not restrictive as we can always add a constant to the unary potentials of a random variable. This modification would only result in the energy of all labelings changing by the same constant. As we shall see, our algorithm as well as its theoretical guarantees are invariant to such changes in the energy function.

\mysection{Advantages of using TMCM}

\begin{table}[h!]
\begin{center}

\begin{tabular}{|r|c|c|c|}
\hline
Labeling & $m=1$ & $m=2$ & $m=3$ \\
\hline
(a) \{1,1,1,1,2,2\} & 1 & 2 & 2 \\
\{1,2,3,4,5,6\} & 3 & 6 & 7 \\
\hline
(b) \{1,1,1,9,9,9\} & 3 & 6 & 9 \\
\{1,1,1,8,8,9\} & 3 & 6 & 9 \\
\hline
(c) \{1,1,1,1,1,7\} & 3 & 3 & 3 \\
\{1,1,1,2,3,4\} & 3 & 5 & 6 \\
\hline
\end{tabular}

\mycaption{\footnotesize \em Clique potential value $\theta_{\bf c}({\bf x}_{\bf c})$ defined by a linear function with $M$ = 3 and $\bf{\omega_c}$ = 1 for various
		values of $m$. Since clique size is 6, $0 \leq m \leq 3$. Pair (a)
		demonstrates why taking the largest convex distances favors smoothness; (b) demonstrates how truncation prevents overpenalization of
		discontinuities; (c) demonstrates how using $m > 1$ can provide some degree of robustness to errors in the definitions of the cliques.
}
\label{table:cliqueExample}
\end{center}
\end{table}

We know show how TMCM encourages smooth labelings, does not overpenalize discontinuities and is robust to erroneous clique definitions.

\myparagraph{\bf Smoothness.} The truncated max-of-convex potentials encourage smooth labelings. In order to illustrate this, let us consider the
example of a clique of six random variables ${\bf X}_{\bf c}$ and a label set ${\bf L}$ of size $10$. We can define a truncated convex distance
using a linear function $d(y) = |y|$ and a truncation factor of $M=3$. Consider pair (a) of labelings shown
in table~\ref{table:cliqueExample}. If the labeling is \{1, 1, 1, 1, 2, 2\} and $M$ = 3, $m$ = 3, $\bf\omega_c$ = 1, $\theta_{\bf c}({\bf x}_{\bf c})$ = min(6-1, 3) + min(5 - 2, 6) + min(4 - 3, 3) = 3 + 3 + 1 = 7. Clearly, the first labeling of this pair is significantly smoother than the second, which is reflected in the value of the
clique potential for all values of $m$. In contrast, if we were to consider the minimum distance among all pairs of labels, both the labelings will
provide a clique potential value of $0$.  
\vspace{-1mm}

\myparagraph{\bf Discontinuities.} Similar to the pairwise case, the use of a truncation factor helps prevent overpenalization of discontinuities. For example,
let us consider the pair (b) of labelings in table~\ref{table:cliqueExample}. In both cases, the six random variables appear to belong to two
{\em groups}, one whose labels are low and one whose labels are high. Without a truncation, such a discontinuity would have been penalized heavily (for example,
8 for $m=1$ for both the labelings). This in turn would discourage the clique to be assigned this labeling even though this type of discontinuity is expected
to occur in natural images. However, with a truncation, the penalty is significantly less (for example, 3 for $m=1$ for both the labelings), which can help
preserve the discontinuities in the labeling obtained via energy minimization.
\vspace{-1mm}

\myparagraph{\bf Robustness.} In order to use a TMCM, we would be required to define the cliques. For example, given an image, we could use a bottom-up oversegmentation
approach to obtain superpixels, and then set all the pixels in a superpixel to belong to a clique. However, oversegmentation can introduce errors since it has
no notion of the specific vision application we are interested in modeling. To add robustness to errors in the clique definitions, we can set $m > 1$.
For example, consider pair (c) of labelings in table~\ref{table:cliqueExample}. The first of these labelings contains a single
random variable with a very high label, which could be due to the fact that the corresponding pixel has been incorrectly grouped in this superpixel. 
As can be seen from the values of the potential, the
presence of such an erroneous pixel in the superpixel is not heavily penalized when we use $m > 1$. For example, when $m=3$ the value of the clique potential for
the first labeling (with an erroneous pixel) is the same as the second labeling (which is a fairly smooth labeling).

\begin{table}
\begin{center}
\setlength{\tabcolsep}{10pt}
\renewcommand{\arraystretch}{1.3}
\begin{tabular}{|c|l|}
\hline
$n$ & Number of random variables \\ 
$h$ & Number of labels \\ 
$\mathcal{V}$ & $\{1, 2, \dots, \mathit{n}\}$ \\
$\mathbf{L}$ &  $\{1, 2, \dots, \mathit{h}\}$ \\ 
$\mathbf{X}$ & Set of random variables $\{\mathit{X}_a, a \in \mathcal{V}\}$ \\ 
$\mathbf{X_c}$ & Set of random variable belonging to a clique $\{X_a, a \in \mathbf{c} \subseteq \mathcal{V}\}$ \\ 
$\mathbf{x_c}$ & Labeling of clique $\mathbf{c}$ \\ 
$\mathbf{p(x_c)}$ & Sorted list of the labels present in $\mathbf{x_c}$ \\ 
$\mathit{I_n}$ & Interval of consecutive labels $[i_n + 1, j_n]$ \\ 
$\mathit{h'}$ & Length of interval, that is, $\mathit{h'} = j_n - i_n$ \\ 
$\Gamma_r$ & Set of intervals $\{[0, r], [r + 1, r + \mathit{L}],\dots, [., h - 1]\}$ \\ 
$\mathit{f}$ & Labeling of the random field ($v_a$ takes the label $l_{f(a)}$) \\
$\mathit{f^*}$ & An optimal (MAP) labeling of the random field \\
$\mathit{\theta}_a(i)$ & Unary potential of assigning label $l_i$ to $v_a$ \\
$\mathbf{\omega_c}$ & Weight for clique $\mathbf{X_c}$ \\ 
$\mathit{c}$ & Size of clique $\mathbf{X_c}$ \\ 
$\mathit{d(.)}$ & Convex function used to define distance between two labels \\
$\mathit{M}$ & Truncation factor \\
$\mathit{\theta}_\mathbf{c}(\mathbf{x_c})$ & Clique potential of assigning label $l_i$ to $v_a$ \\
$\mathit{E(f)}$ & Energy of the labeling $f$ \\
$\mathcal{A}\mathit(f, I_n)$ & \{$\mathbf{X_c} \in \mathcal{C} , f(v) \in I_n \forall v \in \mathbf{X_c}$\} \\
$\mathcal{B}\mathit(f, I_n)$ & \{$\mathbf{X_c} \in \mathcal{C} , \exists X_v, X_w \in \mathbf{X_c} | f(v) \in I_n \land f(w) \notin I_n$ \} \\
\hline
\end{tabular}
\end{center}
\caption{Definitions of various symbols used in the supplementary document}
\end{table}

\mysection{Optimization via Range Expansion}

As TMCM is a generalization of TCM, it follows that the corresponding energy
minimization problem is NP-hard. However, we show that the efficient and accurate
range expansion algorithm can be extended to handle this more general class
of energy functions.

Algorithm~\ref{algo:rangeExpansion} shows the main steps of range expansion. The algorithm starts by assigning
the random variables to an initial label (step 1). For example, all the random variables could be assigned to the label $1$. 
Next, it selects an interval of consecutive labels of size at most $h'$ (steps 3-4), where $h'$ is specified as an input to the
algorithm. We will see later in the section that the value of $h'$ can be chosen to obtain the optimal worst case bound for specific instances of
the TMCM. Next, it minimizes the energy over all the labelings that either allow a random
variable to retain its current label, or choose a new label in the selected interval (step 5). If the energy of the new labeling is lower than
that of the current labeling, then the solution is updated (steps 6-8). This process is repeated for all the intervals
of consecutive labels of size at most $h'$. The entire algorithm stops when the energy cannot be reduced further for any choice of the interval.

\begin{algorithm}
\caption{The range expansion algorithm.}
\begin{algorithmic}[1]
\INPUT Energy function $E(\cdot)$, initial labeling ${\bf x}^0$, interval length $h'$.
\STATE Initialize the output labeling $\hat{\bf x} = {\bf x}^0$.
\REPEAT
\FORALL{$i_m \in [-h'+2,h]$}
\STATE Define an interval of labels ${\bf I} = \{f,\cdots,l\}$ where $f = \max\{i_m,1\}$ and
$l = \min\{i_m+h'-1,h\}$.
\STATE Obtain a new labeling ${\bf x}'$ by solving the following optimization problem:
\begin{eqnarray}
{\bf x}' = && \argmin_{\bf x} E({\bf x}), \nonumber \\
&& \mbox{s.t. } x_a \in {\bf I} \cup \{\hat{x}_a\}, \forall a \in {\cal V}.
\label{eq:rangeMove}
\end{eqnarray}
\IF{$E(\hat{\bf x}) > E({\bf x}')$}
\STATE Update $\hat{\bf x} = {\bf x}'$.
\ENDIF
\ENDFOR
\UNTIL The labeling does not change for any value of $i_m$.
\OUTPUT The labeling $\hat{\bf x}$.
\end{algorithmic}
\label{algo:rangeExpansion}
\end{algorithm}

The crux of the range expansion algorithm is problem~(\ref{eq:rangeMove}), which needs to be solved for any given interval ${\bf I}$ and
current labeling $\hat{\bf x}$. Unfortunately, this problem itself is NP-hard for TMCM. Indeed, when $h' = h$, problem~(\ref{eq:rangeMove}) is
equivalent to the original energy minimization problem. In order to operationalize the range expansion algorithm, we need to devise an
approximate algorithm for problem~(\ref{eq:rangeMove}). We achieve this in two steps. First, we obtain an overestimate of the energy
function $E(\cdot)$, which we denote by $E'(\cdot)$.
The energy function $E'(\cdot)$ is restricted to the labels in the interval ${\bf I}$ together with the labels specified by
the current labeling $\hat{\bf x}$. Second, we minimize the overestimated energy $E'(\cdot)$ over all of its putative
labelings by solving an equivalent $st$-{\sc mincut} problem.
We describe our two-step algorithm in the next two subsections in detail. Specifically, subsection~\ref{subsec:overestimate} describes the
exact form of the energy function $E'(\cdot)$, while subsection~\ref{subsec:graph} describes the construction of the
directed graph over which we solve the $st$-{\sc mincut} problem to obtain the labeling ${\bf x}'$.

\mysubsection{Overestimation of the Energy Function}
\label{subsec:overestimate}

Given an interval ${\bf I} = \{f,\cdots,l\}$ of consecutive labels, and the current labeling $\hat{\bf x}$, we define the new energy function $E'(\cdot)$ over the set of random variables ${\bf X}$. Unlike the original energy function, the label set corresponding to $E'(\cdot)$ is equal to ${\bf L}' = \{0,1,\cdots,h'\}$, where $h'=l - f + 1$.  The label $0$ in the set ${\bf L}'$ corresponds to a random variable retaining its current label, while any other label $i \geq 1$ corresponds to a random variables taking the label ${f + i - 1} \in {\bf I}$. A labeling of the energy function $E'(\cdot)$ is denoted by ${\bf y} \in ({\bf L}')^n$ in order to distinguish it from the labeling corresponding to the original energy function. We say that a labeling ${\bf x} \in {\bf L}^n$ corresponds to the labeling ${\bf y} \in ({\bf L}')^n$ if
\begin{equation}
x_a = 
\begin{cases}
\hat{x}_a & \text{if } y_a = 0, \\
y_a+f-1 & \text{otherwise.}
\end{cases}
\label{eq:labelingMap}
\end{equation}

We define the unary potentials and the clique potentials of the energy function $E'(\cdot)$ as follows.  

\paragraph{Unary Potentials.} The unary potential of a random variable $X_a$ (where $a \in {\cal V}$) being assigned a label $y_a \in {\bf L}'$ is given by the following equation:

\begin{equation}
\theta'_a(y_a) = \left\{
\begin{array}{cl}
\theta_a(\hat{x}_a) + \kappa_a & \mbox{if } y_a = 0\\
\theta_a(y_a+f-1) + \kappa_a & \mbox{otherwise.}
\end{array}
\right.
\end{equation}
In other words, if $y_a = 0$ then the unary potential corresponds to the random variable $X_a$ retaining its current label $\hat{x}_a$, and if $y_a \neq 0$ then the unary potential corresponds to the random variable $X_a$ being assigned the label $y_a+f-1 \in {\bf I}$.  The constant $\kappa_a$ is added to the unary potentials to ensure that they are non-negative, which makes the description of the graph construction in the next subsection simpler.

\paragraph{Clique Potentials.} In order to describe the high-order clique potentials of the new energy function we require a function $\delta_{a,b}: {\bf L}' \times {\bf L}' \rightarrow \mathbb{R}$ for each $(a,b) \in {\cal E}$, which is defined as follows:

\begin{equation}
\delta_{a,b}(y_a,y_b) = \left\{
\begin{array}{cl}
\min\{d(\hat{x}_a-\hat{x}_b),M\} & \mbox{if } y_a = y_b = 0, \\
M+d(y_b-1) & \mbox{if } y_a = 0, y_b \neq 0, \\
M+d(y_a-1) & \mbox{if } y_a \neq 0, y_b = 0, \\
d(y_a-y_b) & \mbox{if } y_a \neq 0, y_b \neq 0.
\end{array}
\right.
\label{eq:submodOverestimate}
\end{equation}
Here, $d(\cdot)$ is the convex function and $M$ is the truncation factor associated with the original energy function $E(\cdot)$. 

\begin{proposition}
\label{prop:submodOver}
$\delta_{a,b}(y_a,y_b)$ is submodular in the sense of label-set encoding used in ~\cite{ishikawapami03}and is an overestimate of the truncated convex distance $\min\{d(x_a-x_b),M\}$ (Proof in Appendix \ref{app:submodOverest}).
\end{proposition}

Given a labeling ${\bf y}_{\bf c} \in ({\bf L'})^c$ of a clique ${\bf c}$ of size $c$, we denote a sorted list of the labels in ${\bf y}_{\bf c}$ as ${\bf p}({\bf y}_{\bf c})$. Furthermore, we denote the indices of the sorted list as ${\bf q}({\bf y}_{\bf c})$. In other words, the random variable corresponding to the $i$-th smallest label (that is, the $i$-th element of the list ${\bf p}({\bf y}_{\bf c})$, which is denoted by $p_i({\bf y}_{\bf c})$) is given by $X_a$ where $a = q_i({\bf y}_{\bf c})$. To avoid clutter, we will drop the argument ${\bf y}_{\bf c}$ from ${\bf p}$ and ${\bf q}$ whenever it is clear from context.  

Using the above definitions, the high-order clique potential for the new energy $E'(\cdot)$ can be concisely specified as

\begin{equation}
\theta'_{\bf c}({\bf y}_{\bf c}) = \omega_{\bf c} \sum_{i=1}^m \delta_{q_i,q_{c-i+1}}(p_i,p_{c-i+1}).
\end{equation}
Hence, the clique potentials in the energy function $E'(\cdot)$ are the sum of the $m$ maximum submodular functions over disjoint pairs of random variables in the clique.

\mysubsection{Graph Construction}
\label{subsec:graph}

\begin{figure*}
\centerline{
\includegraphics[width=0.60\textwidth]{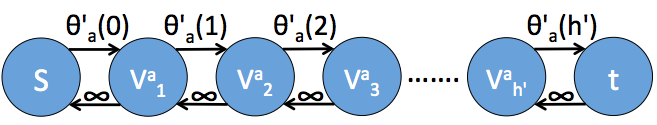}
}
\vspace{3mm}
\mycaption{\footnotesize \em Arcs and their capacities for representing the unary potentials for the random variable $X_a$. According to the labeling defined in equation~(\ref{eq:labeling}), if $x_a = \hat{x}_a$, then the arc $(s,V^a_1)$ will be cut, which will contribute exactly $\theta_a(\hat{x}_a)$ to the capacity of the cut.  If $x_a = s+i-1$ where $i \in \{1,\cdots,h'-1\}$, then the arc $(V^a_i,V^a_{i+1})$ will be cut, which will contribute exactly $\theta_a(s+i-1)$ to the capacity of the cut. If $x_a = l$, then the arc $(V^a_{h'},t)$ will be cut, which will contribute exactly $\theta_a(l)$ to the capacity of the cut. The arcs with infinite capacity ensure that exactly one of the arcs from the set $(s,V^a_1) \cup \{(V^a_i,V^a_{i+1}), i=1,\cdots,h'-1\} \cup (V^a_{h'},t)$ will be part of an $st$-cut with finite capacity, which will guarantee that we are able to obtain a valid labeling.
}
\label{fig:unary}
\end{figure*}

\begin{figure*}[t]
\centering
\hspace{1cm}
\includegraphics[width=0.30\textwidth]{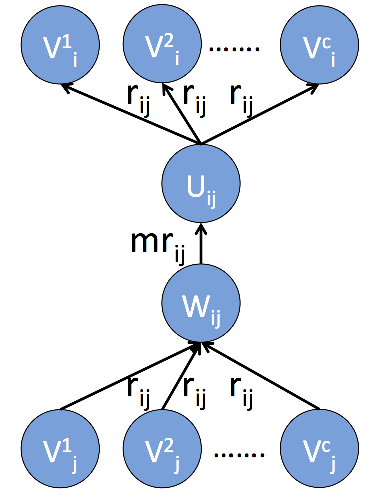}
\hspace{1cm}
\includegraphics[width=0.30\textwidth]{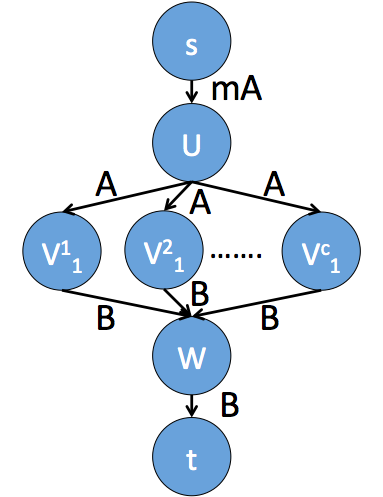}
\mycaption{\footnotesize \em Arcs used to represent the high-order potentials for the clique ${\bf X}_{\bf c} = \{X_1,X_2,\cdots,X_c\}$. {\bf Left.} The term $r_{ij}$ is defined in equation~(\ref{eq:convexCapacity}). The arcs represent the sum of the $m$ maximum convex distance functions over disjoint pairs of random variables when no random variable retains its old label. These arcs are specified only for $i \leq j$ and when either one or both of $i$ and $j$ are not equal to 1. {\bf Right.} The terms $A$ and $B$ are defined in equation~(\ref{eq:submodCapacity}). The arcs represent an overestimation of the clique potential for the case where some or all the random variables retain their old label.}
\label{fig:clique}
\end{figure*}

Our problem is to minimize the energy function $E'(\cdot)$ over all possible labelings ${\bf y} \in ({\bf L'})^n$.
To this end, we convert it into an equivalent $st$-{\sc mincut} problem over a directed graph, which can be solved efficiently if all arc capacities are non-negative~\cite{boykovpami04}.

We construct a directed graph over the set of vertices $\{s,t\} \cup {\bf V} \cup {\bf U} \cup {\bf W}$. The set of vertices ${\bf V}$ model the random variables
${\bf X}$. Specifically, for each random variable $X_a$ we define $h'=l-f+1$ vertices $V^a_i$ where $i \in \{1,\cdots,h'\}$.  The sets ${\bf U}$ and ${\bf W}$ represent {\em auxiliary} vertices, whose role in the graph construction will be explained later when we consider representing the high-order clique potentials. We also define a set of arcs over the vertices, where each arc has a non-negative capacity.  We would like to assign arc capacities such that the $st$-cuts of the directed graph satisfy two properties. First, all the $st$-cuts with a finite capacity should include exactly one arc from the set $(s,V^a_1) \cup \{(V^a_i,V^a_{i+1}), i=1,\cdots,h'-1\} \cup (V^a_{h'},t)$ for each random variable $X_a$. This property would allow us to define a labeling ${\bf y}$ such that

\begin{equation}
\small y_a = \left\{
\begin{array}{cl}
0 & \mbox{if the cut includes the arc } (s,V^a_1) \\
i & \mbox{if the cut includes the arc } (V^a_i,V^a_{i+1}) \\
h' & \mbox{if the cut includes the arc } (V^a_{h'},t). \\
\end{array}
\right.
\label{eq:labeling}
\end{equation}

Second, we would like the energy of the labeling {\bf y} defined above to be as close as possible to the capacity of the $st$-cut. This will allow us to obtain an optimal labeling with respect to the energy function $E'()$  by finding the $st$-{\sc mincut}. We now specify the arcs and their capacities such that they
satisfy the above two properties. We consider two cases: (i) arcs that represent the unary potentials; and (ii) arcs that represent the high-order clique potentials.

\myparagraph{\bf Representing Unary Potentials.}
We will represent the unary potential
of $X_a$ using the arcs specified in Figure~\ref{fig:unary}. Since all the unary potentials are non-negative, it follows that the arc
capacities in Figure~\ref{fig:unary} are also non-negative.

\myparagraph{\bf Representing Clique Potentials.}
Consider a set of random variables ${\bf X}_{\bf c}$ that are used to define a high-order clique potential. Without loss of generality, we assume
${\bf X}_{\bf c} = \{X_1,X_2,\cdots,X_c\}$. In order to represent the potential value for a putative labeling ${\bf x}_{\bf c}$ of the clique, we
introduce two types of arcs, which are depicted in Figure~\ref{fig:clique}. For the arcs shown in Figure~\ref{fig:clique} (left), the capacities are
specified using the term $r_{ij}$ that is defined as follows:
\begin{equation}
\small r_{ij} = 
\begin{cases}
\omega_{\bf c}\frac{\overline{d}(i,j)}{2} & \text{if } i = j\\
\omega_{\bf c}\overline{d}(i,j)   & \text{otherwise.}
\end{cases}
\label{eq:convexCapacity}
\end{equation}
Here, the term $\overline{d}(i,j) = d(i-j+1) + d(i-j-1) - 2d(i-j) \geq 0$ since $d(\cdot)$ is convex, and $\omega_{\bf c} \geq 0$ by definition. It follows that $r_{ij} \geq 0$ for all $i,j \in \{1,\cdots,h'\}$. 
We have the following important lemma which will be helpful in the discussion of graph properties later:

\begin{lemma}
	Graph (a) of figure~\ref{fig:clique} exactly models clique potentials that are proportional to the sum of $m$ maximum convex distance functions over all disjoint pairs of random variables of the clique. (Proof in Appendix \ref{app:graphProp})
\label{lemma:cliqueGraphProof}
\end{lemma}

For the arcs shown in Figure~\ref{fig:clique} (right), the capacities are specified using the terms $A$ and $B$ that are defined as follows:
\begin{equation}
\small A = \omega_{\bf c}M, B = \left(\omega_{\bf c}M-\frac{\theta_{\bf c}(\hat{\bf x}_{\bf c})}{m}\right).
\label{eq:submodCapacity}
\end{equation}
Since $M \geq 0$, and $\theta_{\bf c}(\hat{\bf x}_{\bf c}) \leq \omega_{\bf c}mM$ due to truncation, it follows that $A, B \geq 0$.

\subsubsection{Properties of the Graph}
\label{subsec:graph_properties}

This part describes the properties of the above graph construction which will facilitate the analysis of our algorithm for the truncated max-of-linear models. 

\begin{property}The cost of the $st-$cut exactly represents the sum of the unary potentials associated with the corresponding labeling $f$, that is, $\sum_{v_a \in \bf_v} \theta_{a}(f(a))$.
\end{property}
This property follows from the description of the graph construction for unary potentials.
\begin{property} For $c \in \mathcal{C}$, if for all $X_a \in \mathbf{X_c}$, $f(a) = f_n(a) \notin I_n$ (all variables in a clique retain their old labels), then the cost of the $st-$cut exactly represents the clique potential plus a constant ${\bf \kappa}_{\bf c} = \omega_c m \cdot M - \theta_{\bf c}(\hat{{\bf x}_{\bf c}})$.
\end{property}
Graph (a) of figure~\ref{fig:clique} assigns $0$ cost. In graph (b), the vertices $\{V_1^1, \dots, V_1^c \} \in {\bf V}_{\bf t}$ and hence, the $s-U$ arc of capacity $mA$ is cut. 

\begin{align*}
mA &= \omega_c \cdot m \cdot M \\
&= \theta_{\bf c}(\hat{{\bf x}_{\bf c}}) + \omega_c m \cdot M - \theta_{\bf c}(\hat{{\bf x}_{\bf c}})\\
&= \theta_{\bf c}(\hat{{\bf x}_{\bf c}}) + {\bf \kappa}_{\bf c}
\end{align*}

\begin{property} If all variables in a clique move to a label in the current interval and $|l_c - l_1| \leq M$, then the cost of the $st-$cut exactly represents the clique potential plus a constant ${\bf \kappa}_{\bf c} = \omega_c m \cdot M - \theta_{\bf c}(\hat{{\bf x}_{\bf c}})$.
\end{property}

As proved in lemma~\ref{lemma:cliqueGraphProof}, graph (a) of figure~\ref{fig:clique} assigns exactly $\theta_{\bf c}({\bf x}_{\bf c})$ cost. In graph (b), the vertices $\{V_1^1, \dots, V_1^c \} \in {\bf V}_{\bf s}$ and hence, the $W-t$ arc of capacity $B$ is cut, where  

\begin{equation*}
B = \omega_c \cdot m \cdot M - \theta_{\bf c}(\hat{{\bf x}_{\bf c}}) = {\bf \kappa}_{\bf c}
\end{equation*}

\begin{property} If all variables in a clique move to a label in the current interval and $|l_c - l_1| \geq M$, then the cost of the $st-$cut overestimates the clique potential, being 

\begin{equation*}
\omega_c \sum_{i = 1}^{m} d(p_i({\bf x}_{\bf c} - p_{c - i + 1}({\bf x}_{\bf c})
\end{equation*}

plus a constant ${\bf \kappa}_{\bf c} = \omega_c m \cdot M - \theta_{\bf c}(\hat{{\bf x}_{\bf c}})$.

\end{property}

\begin{property} If $k ( < c)$ variables in a clique retains their old labels, then the cost of the $st-$cut overestimates the clique potential, being.
\end{property}

\begin{equation*}
\omega_c \sum_{i = 1}^{k} d(p_{c - i + 1}({\bf x}_{\bf c}) - i_n - 1) + \omega_c \cdot m \cdot M 
\end{equation*}

plus a constant ${\bf \kappa}_{\bf c} = \omega_c m \cdot M - \theta_{\bf c}(\hat{{\bf x}_{\bf c}})$.

The following proposition follows from the properties of the graph:
\begin{proposition}
\label{prop:graphCut}
Given a cut that partitions the vertices ${\bf V}$ into two disjoint sets ${\bf V}_s$ and ${\bf V}_t$, and the corresponding
labeling ${\bf y}$ defined in equation~(\ref{eq:labeling}), the capacity of the cut is equal to the energy $E'({\bf y})$
up to a constant.
\end{proposition}

\paragraph{Energy Minimization.} The above proposition implies that the labeling ${\bf y}'$ corresponding to the $st$-{\sc mincut} minimizes the energy $E'(\cdot)$ over all possible labelings ${\bf y} \in ({\bf L}')^n$. Since all the arc capacities in the graph are non-negative, the labeling ${\bf y}'$ can be computed efficiently by solving the $st$-{\sc mincut} problem on the directed graph defined above. Once the labeling ${\bf y}'$ is computed, we find an approximate solution ${\bf x}'$ to problem~(\ref{eq:rangeMove}) using equation~(\ref{eq:labelingMap}).

\mysubsection{Multiplicative Bounds}
\label{subsec:bounds}
We obtain multiplicative bounds for our algorithm by making use of the fact that our algorithm terminates only when the energy cannot be reduced for any interval $I_n$. This implies that our algorithm has found the local minimum of the large neigbourhood defined by the intervals. We first obtain a lower bound on the reduction in energy for a given interval by making use of the graph properties. Since the final labeling $f$ is a local minimum over the intervals, this lower bound  is non-positive when the algorithm completes. This fact enables us to get an upper bound on the energy of the final labeling. 

We first need to introduce some notation. Let $f_n$ denote the labeling after the $n$-th iteration and $E(f_n)$ denote the corresponding energy. Also, $f^*$ denotes optimal labeling of the MRF.

Let $r \in [0, h'-1]$ be a uniformly distributed random integer and $h'$ be the length of the interval. Using $r$ we define the following set of intervals
\begin{equation*}
	\Gamma_r = \left\{[0, r], [r + 1, r + h'], [r + h' + 1, r + 2h'], ..., [., h -1]\right\}
\end{equation*}
where $h$ is the total number of labels.

Let \textbf{X}$(f^*, I_{n})$ contain all the random variables that take an optimal labeling in $I_{n}$, $\mathcal{A}\mathit(f, I_n)$ be the set of all cliques for which all variables take optimum label in the interval $I_{n}$ and $\mathcal{B}\mathit(f, I_n)$ be the set of all cliques for which at least one, but not all, variable takes optimum label in the interval $I_{n}$ (that is, at least one but not all variables retains old label).

In order to make the analysis more readable, the following shorthand notation is introduced:

\begin{itemize}
	\item We denote $\mathbf{\omega_c} \max_{a, b \in \mathbf{X_c}} d(f^*(a) - f^*(b))$ as $t_c^n$
	\item We denote $\mathbf{\omega_c} \max_{a \in \mathbf{X_c}} d(f^*(a) - (i_n + 1)) + \mathbf{\omega_c}\mathit{M} $ as $s_c^n$
\end{itemize}

The following lemma establishes a lower bound in the reduction in energy for any given interval:

\begin{lemma}
At an iteration of our algorithm, given the current labeling $f_n$ and an interval $I_n = [i_n + 1, j_n]$, the new labeling $f_{n+1}$ obtained by solving the st-mincut problem reduces the energy by at least the following:

\begin{equation*}
	\begin{split}
		& \sum_{X_a \in {\bf X}(f^*, I_{n})} \theta_{a}(f_n(a)) + \sum_{{\bf X_c} \in \mathcal{A}\mathit(f, I_n) \cup \mathcal{B}\mathit(f, I_n)} \theta_{\bf c}({\bf x_c}) \\
		&	  - \left( \sum_{X_a \in {\bf X}(f^*, I_{n})} \theta_{a}(f^*(a)) + \sum_{{\bf X_c} \in \mathcal{A}\mathit(f, I_n)} t_c^n 
	 +  \sum_{{\bf X_c} \in \mathcal{B}\mathit(f, I_n)} s_c^n \right) 
	 \end{split}
\end{equation*}
\label{lemma:reductionlowerBound}
(Proof in Appendix \ref{app:lemmaProofreduction})
\end{lemma}
The following equation can be deduced from the above definitions:

\begin{equation} \label{unary_sum}
	\sum_{X_a \in \mathbf{X}} \theta_a(f^*(a)) = \sum_{I_n \in \mathcal{I}_r} \sum_{X_a \in \mathbf{X}(f^*, I_{n})} \theta_{a}(f^*(a))
\end{equation}

since $f^*(a)$ belongs to exactly one interval in $I_r$ for all $X_a$. 

For the final labeling $f$ of the range expansion algorithm, the term in lemma~\ref{lemma:reductionlowerBound} should be non-positive for all intervals $I_n$ because $f$ is a local optimum. Hence,

\begin{equation*}
	\begin{split}
		&\sum_{X_a \in {\bf X}(f^*, I_{n})} \theta_{a}(f(a)) + \sum_{{\bf X_c} \in \mathcal{A}\mathit(f, I_n) \cup \mathcal{B}\mathit(f, I_n)} \theta_{\bf c}({\bf x_c})\\
	 & \leq \left( \sum_{X_a \in {\bf X}(f^*, I_{n})} \theta_{a}(f^*(a)) + \sum_{{\bf X_c} \in \mathcal{A}\mathit(f, I_n)} t_c^n 
	 +  \sum_{{\bf X_c} \in \mathcal{B}\mathit(f, I_n)} s_c^n \right), \forall I_{n} 
	\end{split}
\end{equation*}

We sum the above inequality over all $I_n \in \Gamma_r$. The summation of the LHS is at least $E(f)$. Also, using \eqref{unary_sum}, the summation of the above inequality can be written as:

\begin{equation*}
E(f) \leq \sum_{X_a \in {\bf X}} \theta_{a}(f^*(a)) + \sum_{I_n \in \Gamma_r}\left(\sum_{{\bf X_c} \in \mathcal{A}\mathit(f, I_n)} t_c^n + \sum_{{\bf X_c} \in \mathcal{B}\mathit(f, I_n)} s_c^n \right)
\end{equation*}

We now take the expectation of the above inequality over the uniformly distributed random integer $r \in [0, h' - 1]$. The LHS of the inequality and the first term on the RHS (that is, $\sum \theta_a(f^*(a)))$ are constants with respect to $r$. Hence, we get

\begin{equation}
E(f) \leq \sum_{X_a \in {\bf X}} \theta_{a}(f^*(a)) + \frac{1}{h'}\sum_{r}\sum_{I_n \in \Gamma_r}\left(\sum_{{\bf X_c} \in \mathcal{A}\mathit(f, I_n)} t_c^n + \sum_{{\bf X_c} \in \mathcal{B}\mathit(f, I_n)} s_c^n \right)
\label{eq:upperboundfinalenergy}
\end{equation} 

For linear distance function $d(.)$, we have the following lemma:

\begin{lemma}
When $d(.)$ is linear, that is, $d(x) = |x|$, the following inequality holds true:
\begin{equation}
\begin{split}
& \frac{1}{h'}\sum_{r}\sum_{I_n \in \Gamma_r}\left(\sum_{{\bf X_c} \in \mathcal{A}\mathit(f, I_n)} t_c^n + \sum_{{\bf X_c} \in \mathcal{B}\mathit(f, I_n)} s_c^n \right) \\
& \leq  max \left\{ \frac{c}{2}\left(2 + \frac{h'}{M}\right), \left(2 + \frac{2M}{h'}\right) \right\} \sum_{{\bf c} \in \mathcal{C}} \theta_{\bf c}(\bf {x_c})
\end{split}
\end{equation}
where $\mathit{c}$ is the largest clique in the random field (Proof in Appendix \ref{app:linearIneqProof}).
\label{lemma:linearIneq}
\end{lemma}

Using lemma \ref{lemma:linearIneq} in inequality \ref{eq:upperboundfinalenergy}, we obtain the multiplicative bound for max-of-linear models for $m = 1$:

\begin{proposition}
\label{prop:linearBound}
The range expansion algorithm with $h' = M$ has a multiplicative bound of $O(C)$ for truncated max-of-linear model when $m=1$. The
term $C$ equals the size of the largest clique. Hence, if ${\bf x}^*$ is a labeling with
minimum energy and $\hat{\bf x}$ is the labeling estimated by range expansion algorithm then
\begin{equation}
\small
\sum_{a \in {\cal V}} \theta_a(\hat{x}_a) + \sum_{{\bf c} \in {\cal C}} \theta_{\bf c}(\hat{\bf x}_{\bf c}) \leq \nonumber 
\sum_{a \in {\cal V}} \theta_a({x}^*_a) + O(C) \sum_{{\bf c} \in {\cal C}} \theta_{\bf c}({\bf x}^*_{\bf c}).
\label{eq:linear_bound}
\end{equation}
The above inequality holds for arbitrary set of unary potentials and non-negative clique weights (Proof in Appendix \ref{app:linearBoundProof}).
\end{proposition}

We now state the generalization for Proposition \ref{prop:linearBound} for any given $m$:

\begin{proposition}
\label{prop:mlinearBound}
The range expansion algorithm with $h' = M$ has a multiplicative bound of $O(m \cdot C)$ for truncated max-of-linear model for any general value of $m$. The
term $C$ equals the size of the largest clique. Hence, if ${\bf x}^*$ is a labeling with
minimum energy and $\hat{\bf x}$ is the labeling estimated by range expansion algorithm then
\begin{equation}
\small
\sum_{a \in {\cal V}} \theta_a(\hat{x}_a) + \sum_{{\bf c} \in {\cal C}} \theta_{\bf c}(\hat{\bf x}_{\bf c}) \leq \nonumber 
\sum_{a \in {\cal V}} \theta_a({x}^*_a) + O(C) \sum_{{\bf c} \in {\cal C}} \theta_{\bf c}({\bf x}^*_{\bf c}).
\label{eq:linear_bound}
\end{equation}
The above inequality holds for arbitrary set of unary potentials and non-negative clique weights (Proof in Appendix \ref{app:mlinearBoundProof}).
\end{proposition}

Note that for $m = 1$, the bound of the move making algorithm for parsimonious labeling (which is our baseline) is $\left(\frac{r}{r-1}\right) (C - 1) O(\log h)\min(C, h)$ where $C$ is the size of the largest clique and $h$ is the number of labels~\cite{dokaniaiccv15}. Our algorithm gives a better bound of $O(C)$ and does not depend on the number of labels.

Similar to the case of max-of-linear models, by making use of Theorem 4 of \cite{chekurisoda01}, we obtain the bound for max-of-quadratic models for $m = 1$:

\begin{proposition}
\label{prop:quadraticBound}
The range expansion algorithm with $h' = \sqrt{M}$ has a multiplicative bound of $O(C\sqrt{M})$ for the truncated max-of-quadratic model when $m=1$.
\end{proposition}

\mysection{Experiments} 
\label{subsec:experiments}

To demonstrate the efficacy of our algorithm, we test it on the two problems of image inpainting and denoising, and stereo matching. The final labeling energy and convergence time are used as evaluation criteria. We used the parsimonious labeling algorithm of Dokania \textit{et al.}~\cite{dokaniaiccv15} and the move-making algorithm for the co-occurrence based energy function of Ladicky \textit{et al.}~\cite{ladickyeccv10} as baselines. For comparison, we restrict ourselves to max-of-linear models and $m$ = 1, as the available code for the baselines can only handle this special case. For completeness, we report the results of our range expansion algorithm for other cases of TMCM and on synthetic data as well. 

\mysubsection{Synthetic Data}

\begin{figure*}[t]
\centerline{
\includegraphics[keepaspectratio = true, width =0.9\textwidth]{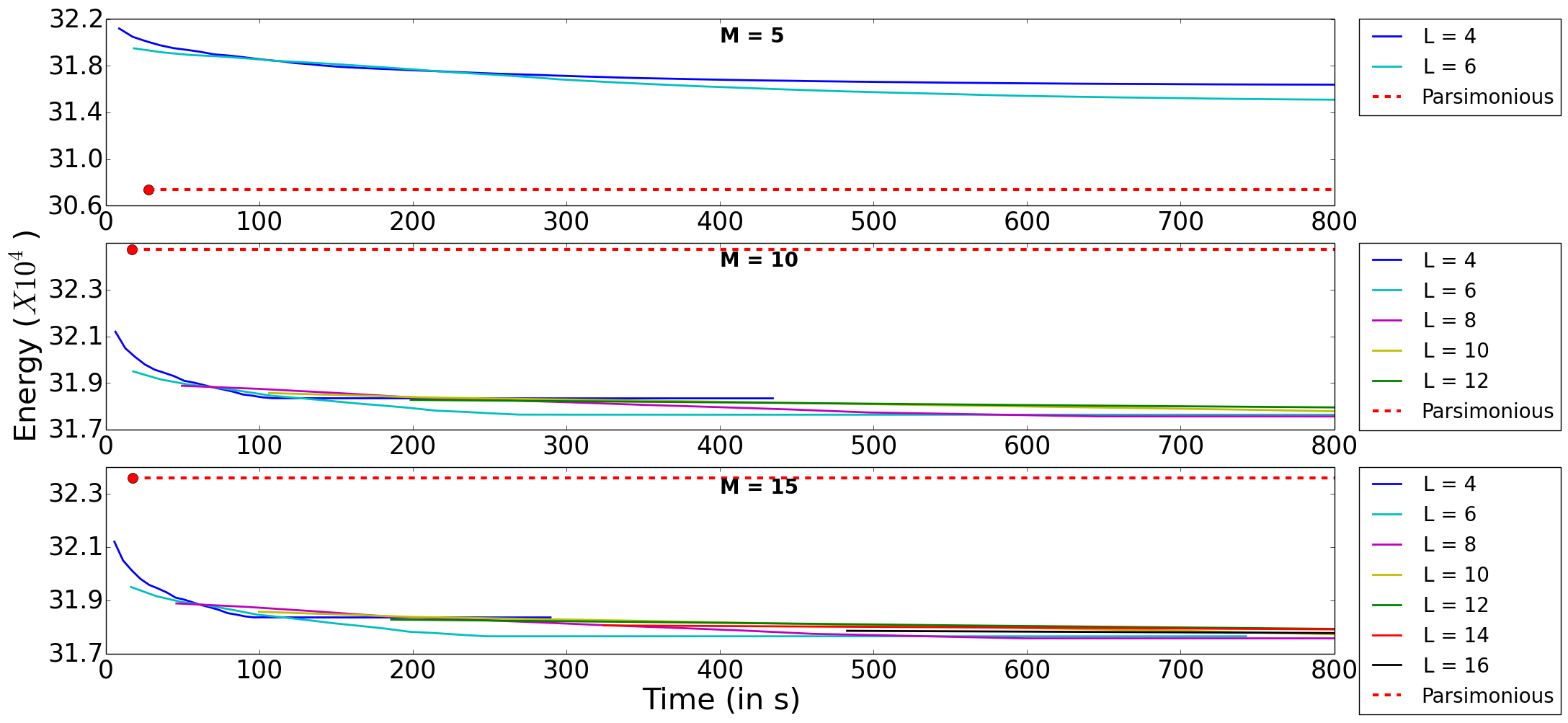}
}
\vspace{3mm}
\mycaption{\footnotesize \em Results for synthetic data using truncated linear distance function. The plots show the variation of energy versus time, averaged over 50 lattices using $\omega_c = 5$. We use truncation factors as $M$ = 5, 10 and 15  and $m$ = 1, and for each we vary interval lengths for our algorithm. Parsimonious labeling performs well for $M$ = 5, but our approach outperforms for higher values of $M$. Red dot indicates convergence of parsimonious labeling and dotted line indicates extrapolation.}
\label{fig:max-of-linear_synthetic}
\end{figure*}

\begin{figure*}
\centerline{
\includegraphics[keepaspectratio = true, width =1.1\textwidth]{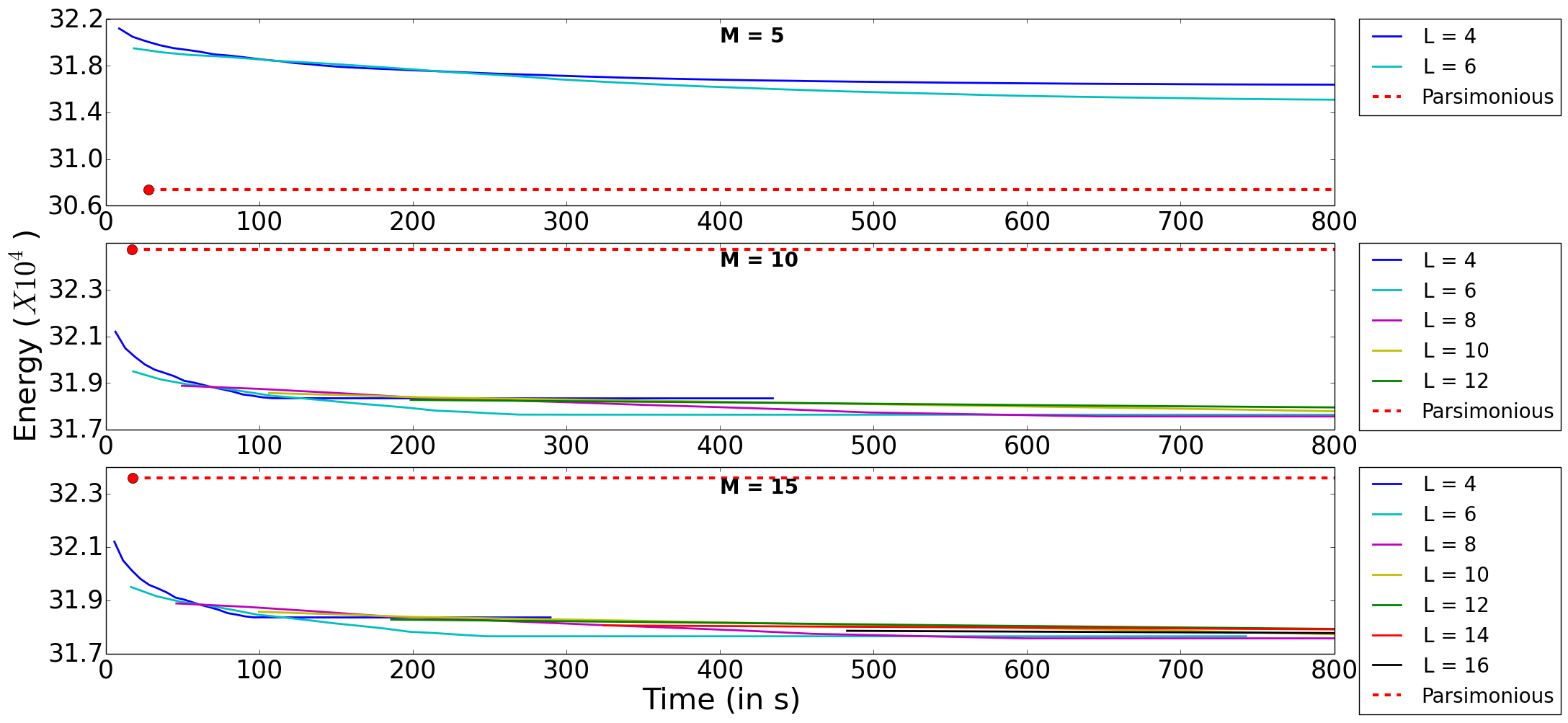}
}
\vspace{3mm}
\mycaption{\footnotesize \em Results for synthetic data using truncated linear distance function. The plots show the variation of energy versus time, averaged over 50 lattices using $\omega_c = 5$. We use truncation factors as $M$ = 5, 10 and 15  and $m$ = 1, and for each we vary interval lengths for our algorithm. This plot is the same as shown in the paper, but we include it here for the sake of comparison. Red dot indicates convergence of parsimonious labeling algorithm and dotted line indicates extrapolation.}
\label{fig:linear_weight5}
\end{figure*}

\begin{figure*}
\centerline{
\includegraphics[keepaspectratio = true, width =1.1\textwidth]{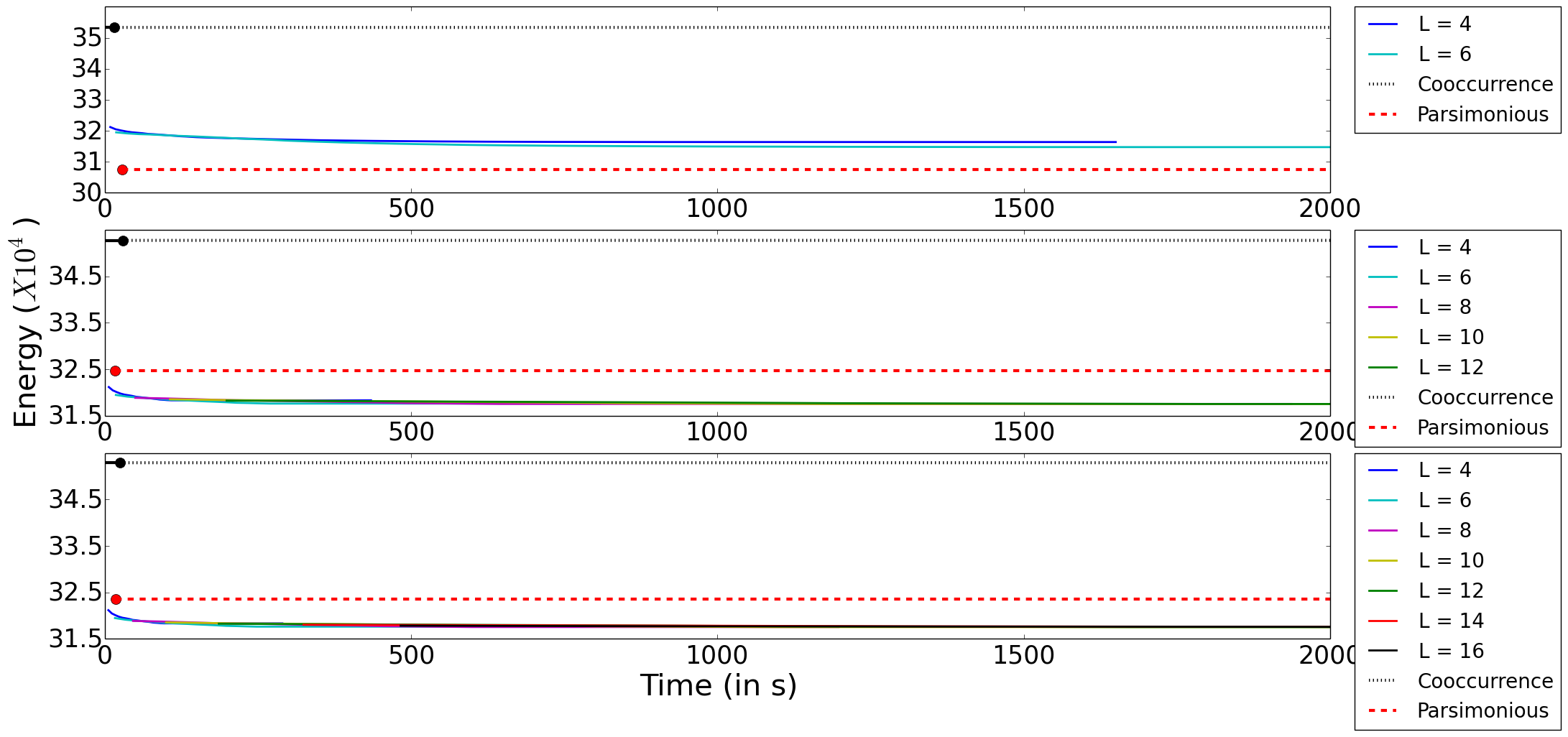}
}
\vspace{3mm}
\mycaption{\footnotesize \em Results for synthetic data using truncated linear distance function. The plots show the variation of energy versus time, averaged over 50 lattices using $\omega_c = 5$. We use truncation factors as $M$ = 5, 10 and 15  and $m$ = 1, and for each we vary interval lengths for our algorithm. This plot corresponds to the same experiment as mentioned in the paper, but with results for co-occurrence included. Red and black dots indicate convergence of respective algorithms and dotted line indicates extrapolation.}
\label{fig:linear_weight5_cooc}
\end{figure*}

\begin{figure*}
\centerline{
\includegraphics[keepaspectratio = true, width =1.1\textwidth]{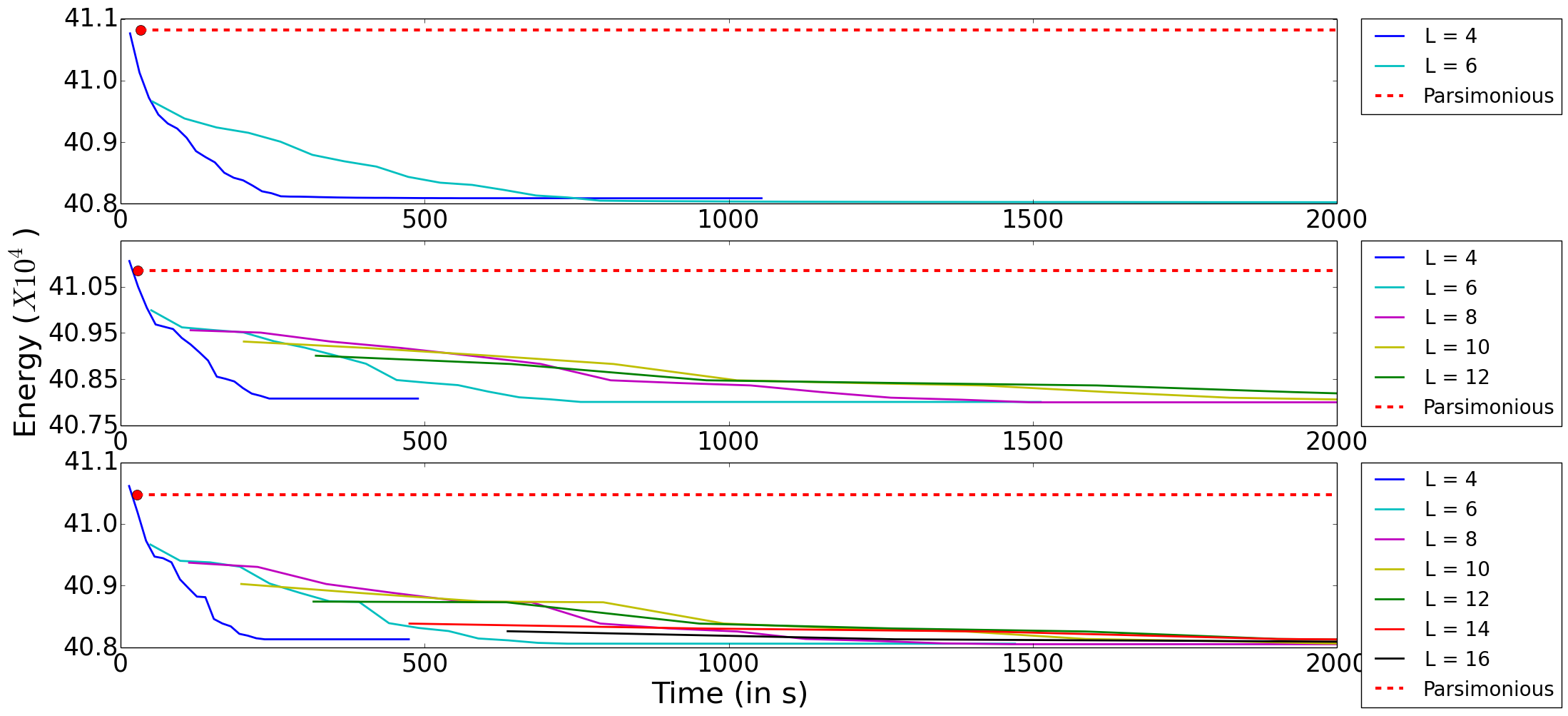}
}
\vspace{3mm}
\mycaption{\footnotesize \em Results for synthetic data using truncated linear distance function. The plots show the variation of energy versus time, averaged over 50 lattices using $\omega_c = 10$. We use truncation factors as $M$ = 5, 10 and 15  and $m$ = 1, and for each we vary interval lengths for our algorithm. Parsimonious labeling performs well for $M$ = 5, but our approach outperforms for higher values of $M$. Red dot indicates convergence of parsimonious labeling and dotted line indicates extrapolation.}
\label{fig:linear_weight10}
\end{figure*}

\begin{figure*}
\centerline{
\includegraphics[keepaspectratio = true, width =1.1\textwidth]{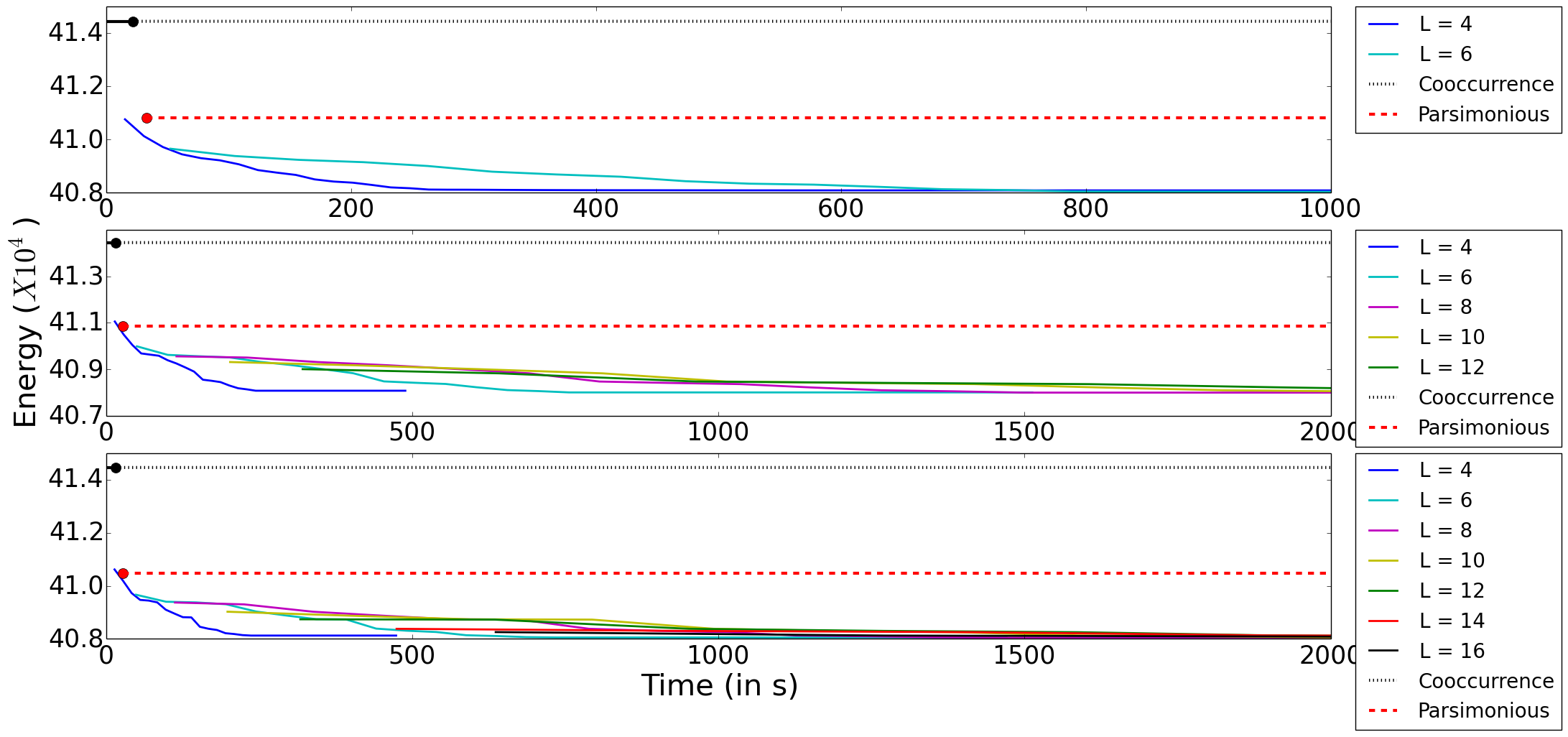}
}
\vspace{3mm}
\mycaption{\footnotesize \em Results for synthetic data using truncated linear distance function. The plots show the variation of energy versus time, averaged over 50 lattices using $\omega_c = 5$. We use truncation factors as $M$ = 5, 10 and 15  and $m$ = 1, and for each we vary interval lengths for our algorithm. This plot corresponds to the same experiment as in figure ~\ref{fig:linear_weight10}, but with results for co-occurrence included. Red and black dots indicate convergence of respective algorithms and dotted line indicates extrapolation.}
\label{fig:linear_weight10_cooc}
\end{figure*}

\begin{figure*}
\centerline{
\includegraphics[keepaspectratio = true, width = 1.1\textwidth]{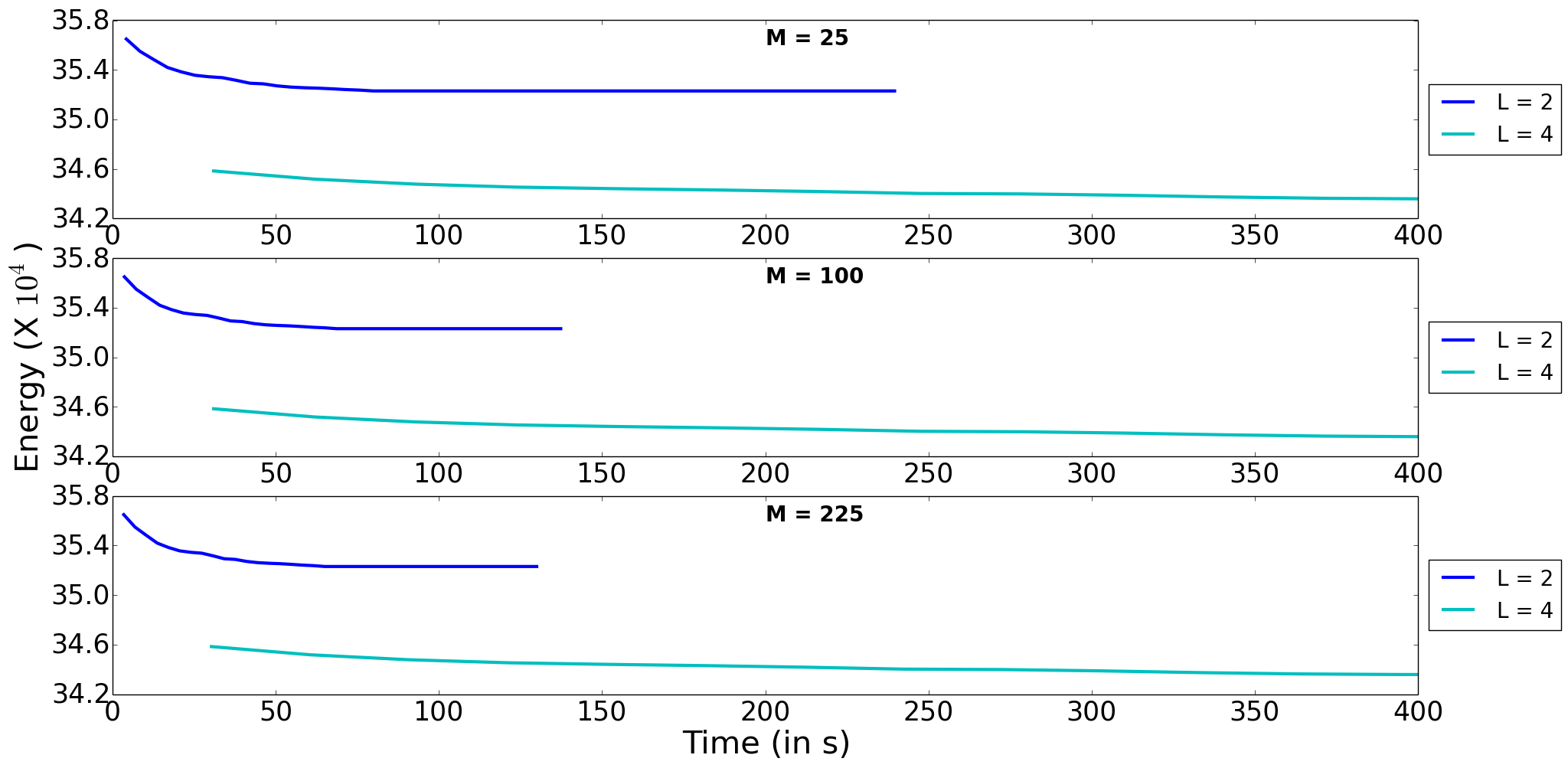}
}
\vspace{3mm}
\mycaption{\footnotesize \em Results for synthetic data using truncated quadratic distance function. The plots show the variation of energy versus time, averaged over 50 lattices using $\omega_c = 3$. We use $M$ = 25, 100 and 225 and $m$ = 1, and for each we vary interval lengths for our algorithm.}
\label{fig:max-of-quadratic_synthetic}
\end{figure*}

\begin{figure*}
\centerline{
\includegraphics[keepaspectratio = true, width = 1.1\textwidth]{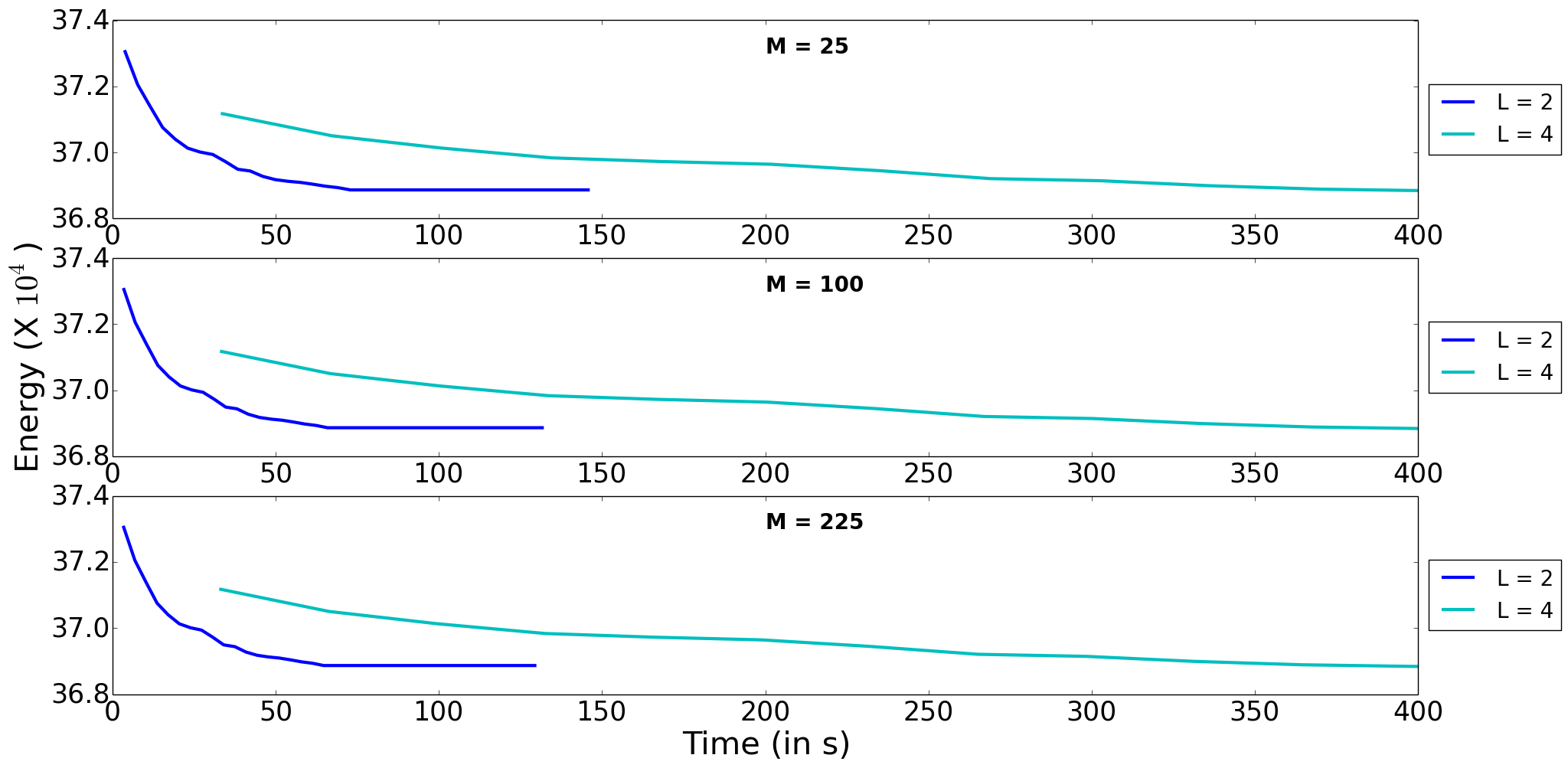}
}
\vspace{3mm}
\mycaption{\footnotesize \em Results for synthetic data using truncated quadratic distance function. The plots show the variation of energy versus time, averaged over 50 lattices using $\omega_c = 5$. We use $M$ = 25, 100 and 225, and for each we vary interval lengths for our algorithm.}
\label{fig:quadratic_weight5}
\end{figure*}

\myparagraph{\bf Data.} We generate lattices of size 100 $\times$ 100, where each lattice point represents a variable taking one of 20 labels. The cliques are defined as all 10 $\times$ 10 subwindows. The unary potentials are uniformly sampled in the range [1, 100]. The truncation factors for linear case are $M \in \{5, 10, 15\}$ and for quadratic case $M \in \{25, 100, 225\}$. In each energy function, all cliques were assumed to have the same clique weight, being 5 and 10 for linear model and 3 and 5 for quadratic model. $m$ varies in the range $\{1, 3, 5\}$.

\myparagraph{\bf Method.} For each energy function obtained by a particular setting of the above parameters, we vary the interval length up to $M$ + 1. We run~\cite{dokaniaiccv15} and~\cite{ladickyeccv10} only for linear distance function and $m$ = 1 and repeat the experiments for 50 randomly generated unaries for linear and quadratic cases. 

\myparagraph{\bf Results.} The plots in Figure~\ref{fig:linear_weight5_cooc} and ~\ref{fig:linear_weight5} show the average energy as a function of average time for our algorithm and the baselines using weight as 5 for max-of-linear models, with and without the results of ~\cite{ladickyeccv10} respectively. We see that the energy values of  ~\cite{ladickyeccv10} are much higher as compared to our algorithm. Our algorithm gives lower energy solutions than parsimonious labeling method except when the truncation factor is low ($M$ = 5). The plots in Figures~\ref{fig:linear_weight10_cooc} and ~\ref{fig:linear_weight10} shows the results for max-of-linear models using weight as 10 with and without the cooccurrence results respectively. Our algorithm gives better results than the baselines for all $M$. Both the baselines converge faster than our method. In practice, intervals smaller than the optimum size give almost equally good results and converge faster. Figure~\ref{fig:quadratic_weight5} shows the plot for max-of-quadratic cases using weight as 5.

\mysubsection{Image Inpainting and Denoising}

\begin{figure}
	\centering
\begin{tabular}{ccc}
	\includegraphics[scale = 0.60]{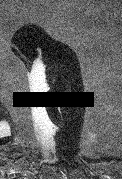} &
	\includegraphics[scale = 0.60]{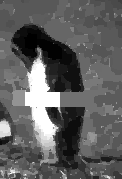} &
	\includegraphics[scale = 0.60]{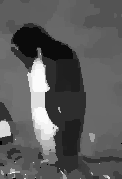}\\
	\scriptsize{(a) Penguin input} & \scriptsize{(b) Cooccurrence} & \scriptsize{(c) Parsimonious} \\ 
	\scriptsize(Energy, Time (s)) & \scriptsize(14735411, 237) & \scriptsize(12585846, 456) \\ 
	\includegraphics[scale = 0.60]{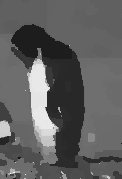} &
	\includegraphics[scale = 0.60]{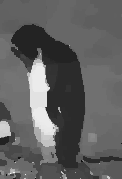} &
	\includegraphics[scale = 0.60]{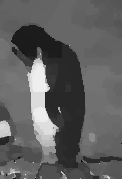} \\ 
	\scriptsize{(d) $m$ = 1, $h'$ = 5} & \scriptsize{(e) $m$ = 1, $h'$ = 10} & \scriptsize{(f) $m$ = 1, $h'$ = 20} \\ 
	\scriptsize(12541999, 1694) & \scriptsize(123598999, 2633) & \scriptsize(123018999, 3963) \\
	\includegraphics[scale = 0.60]{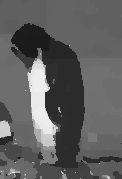} &
	\includegraphics[scale = 0.60]{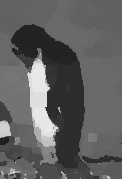} &
	\includegraphics[scale = 0.60]{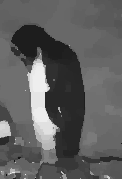} \\ 
	\scriptsize{(g) $m$ = 3, $h'$ = 5} & \scriptsize{(h) $m$ = 3, $h'$ = 10} & \scriptsize{(i) $m$ = 3, $h'$ = 20} \\ 
	\scriptsize(126481999, 1379) & \scriptsize(125784999, 2499) & \scriptsize(124044999, 5018) \\
	\includegraphics[scale = 0.60]{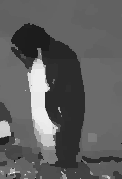} &
	\includegraphics[scale = 0.60]{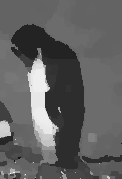} &
	\includegraphics[scale = 0.60]{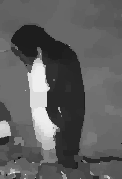} \\ 
	\scriptsize{(j) $m$ = 5, $h'$ = 5} & \scriptsize{(k) $m$ = 5, $h'$ = 10} & \scriptsize{(l) $m$ = 5, $h'$ = 20} \\ 
	\scriptsize(127329999, 1357) & \scriptsize(125284999, 2367) & \scriptsize(124501999, 5706) \\
\end{tabular}
\mycaption{\footnotesize \em Image inpainting results for `penguin'. Note that comparison with (b) and (c) makes sense only for $m$ = 1. Also, we restricted our experiments to smaller (and suboptimal) $h'$ due to computational issues.}
\label{fig:inpainting_results}
\end{figure}

\begin{figure}
	\centering
\begin{tabular}{ccc}
	\includegraphics[scale = 0.40]{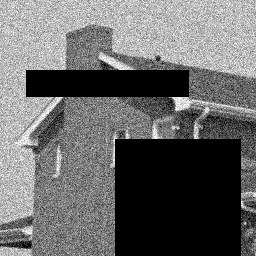} &
	\includegraphics[scale = 0.40]{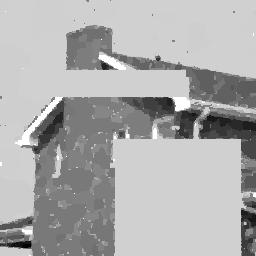} &
	\includegraphics[scale = 0.40]{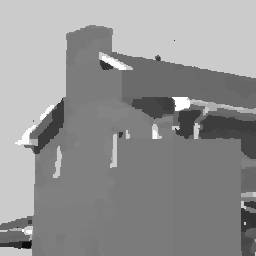}\\
	\scriptsize{(a) Penguin input} & \scriptsize{(b) Cooccurrence} & \scriptsize{(c) Parsimonious} \\ 
	\scriptsize(Energy, Time (s)) & \scriptsize(42018464, 486) & \scriptsize(37349032, 12024) \\ 
	\includegraphics[scale = 0.40]{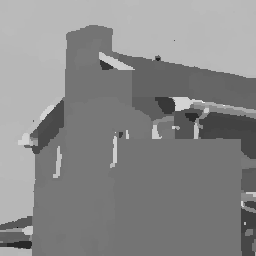} &
	\includegraphics[scale = 0.40]{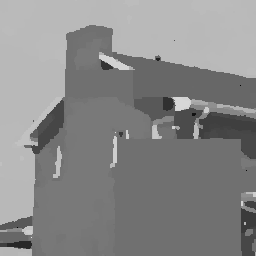} &
	\includegraphics[scale = 0.40]{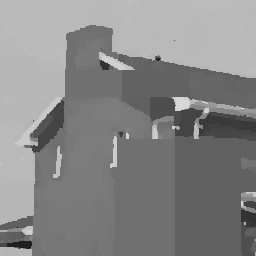} \\ 
	\scriptsize{(d) $m$ = 1, $h'$ = 5} & \scriptsize{(e) $m$ = 1, $h'$ = 10} & \scriptsize{(f) $m$ = 1, $h'$ = 20} \\ 
	\scriptsize(37196999, 8084) & \scriptsize(370873999, 11356) & \scriptsize(369035999, 22752) \\
	\includegraphics[scale = 0.40]{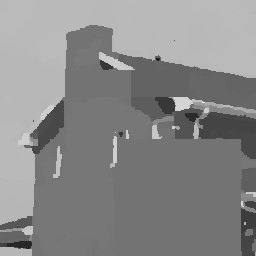} &
	\includegraphics[scale = 0.40]{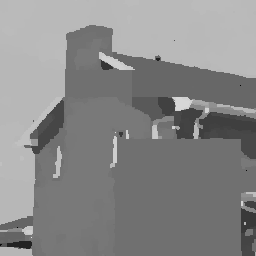} &
	\includegraphics[scale = 0.40]{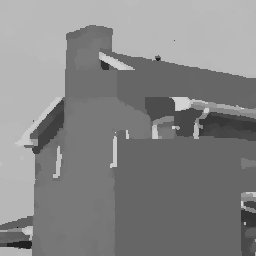} \\ 
	\scriptsize{(g) $m$ = 3, $h'$ = 5} & \scriptsize{(h) $m$ = 3, $h'$ = 10} & \scriptsize{(i) $m$ = 3, $h'$ = 20} \\ 
	\scriptsize(373067999, 7744) & \scriptsize(37115999, 9547) & \scriptsize(370936999, 23261) \\
	\includegraphics[scale = 0.40]{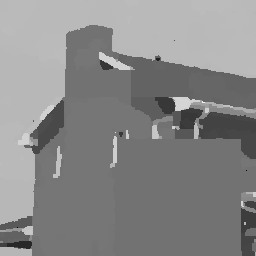} &
	\includegraphics[scale = 0.40]{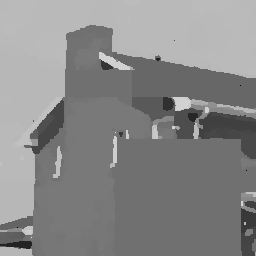} &
	\includegraphics[scale = 0.40]{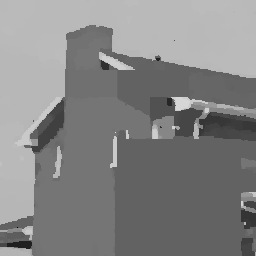} \\ 
	\scriptsize{(j) $m$ = 5, $h'$ = 5} & \scriptsize{(k) $m$ = 5, $h'$ = 10} & \scriptsize{(l) $m$ = 5, $h'$ = 20} \\ 
	\scriptsize(374691999, 6476) & \scriptsize(372811999, 9949) & \scriptsize(375026999, 22985) \\
\end{tabular}
\vspace{2mm}
\mycaption{\footnotesize \em Image inpainting results for `house'. Note that comparison with (b) and (c) makes sense only for $m$ = 1. Also, we restricted our experiments to smaller (and suboptimal) $h'$ due to computational issues.}
\label{fig:inpainting_results}
\end{figure}

\myparagraph{\bf Data.} Given an image with noise and obscured/inpainted regions (regions with missing pixels), the task is to denoise it and fill the obscured regions in a way that is consistent with the surrounding regions. The images `house' and `penguin' from the Middlebury data set were used for the experiments. Since the images are grayscale, they have 256 labels in the interval [0, 255], each representing an intensity value. The unary potential for each pixel corresponding to a particular label equals the squared difference between the label and the intensity value at the pixel. We use high-order cliques as the super-pixels obtained using the mean-shift method~\cite{comaniciu2002mean}. The parameters $\omega_c$, $M$ and $m$ are varied to give different truncated max-of-linear energy functions.

\vspace{-1mm}
\myparagraph{\bf Method.} For each parameter setting of $\omega_c$, $M$ and $m$, we vary the interval lengths for our algorithm and make a comparison with the baselines.

\vspace{-1mm}

\myparagraph{\bf Results.} Results for $\omega_c$ = 40, $M$ = 40, and $m$ = 1, 3 and 5 for `penguin' and $\omega_c$ = 50, $M$ = 50, and $m$ = 1, 3 and 5 for `house' are shown in Figure~\ref{fig:inpainting_results} for varying interval lengths $h' = \{5, 10, 20\}$. Our algorithm consistently gives lower energy labeling as compared to both~\cite{dokaniaiccv15} and~\cite{ladickyeccv10} even for small $h'$. For `penguin', our algorithm gives cleaner denoised image, preserving edges and details. On the other hand, both~\cite{dokaniaiccv15} and~\cite{ladickyeccv10} exhibit significant blocking effect. Moreover, the output is more natural for $m$ = 3 as compared to $m$ = 1. Even for `house', our output looks more visually appealing as compared to baselines.

\clearpage
\mysubsection{Stereo Matching}
\myparagraph{\bf Data.} In the stereo matching problem, we have two rectified images of the same scene from two cameras set slightly apart. We are required to estimate the horizontal disparity between a pixel in the right camera image from the corresponding pixel in the left camera. We use `tsukuba' and `teddy' data sets from the Middlebury stereo collection for our experiments. In each case, we have a pair of RGB images and ground truth disparities. We assume the unary potentials to be the $L1$-norm of the difference in RGB values of the left and right image pixels. There are 16 labels and 60 labels for `tsukuba' and `teddy' respectively. The high-order cliques are super-pixels obtained using mean-shift method~\cite{comaniciu2002mean}. The parameters $\omega_c$, $M$ and $m$ are varied to give different truncated max-of-linear energy functions.

\myparagraph{\bf Method.} For each parameter setting of $\omega_c$, $M$ and $m$, we vary the interval lengths for our algorithm and make a comparison with the baselines.

\begin{figure}
\centering
\begin{tabular}{cccc}
	\includegraphics[scale = 0.20]{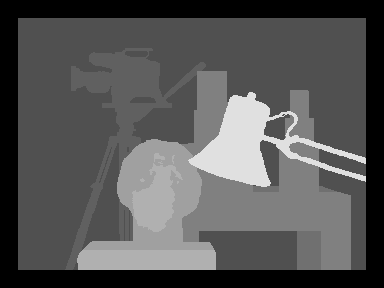} &
	\includegraphics[scale = 0.15]{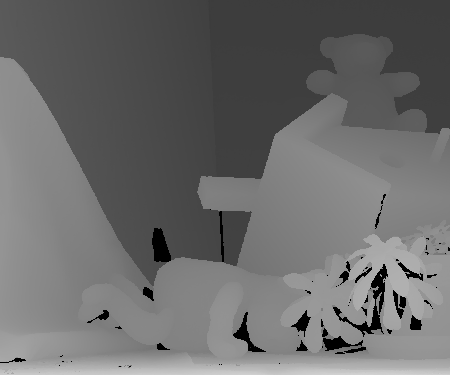} &
	\includegraphics[scale = 0.15]{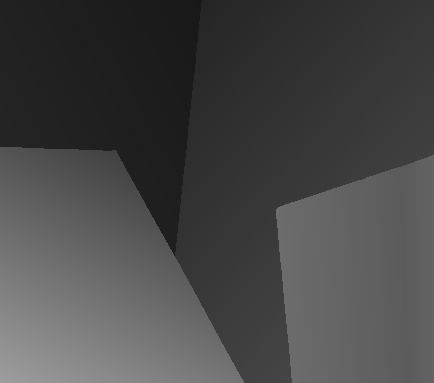} &
	\includegraphics[scale = 0.15]{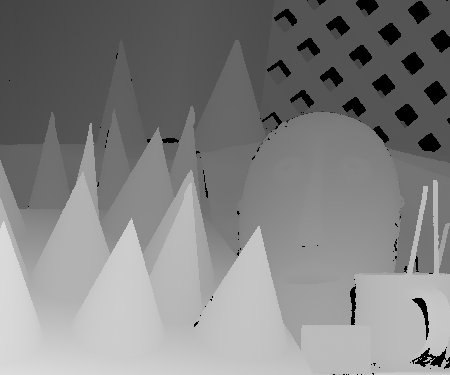} \\
	\scriptsize(1a) Ground truth & \scriptsize(2a) Ground truth & \scriptsize(3a) Ground truth & \scriptsize(4a) Ground truth \\
	\scriptsize(Energy, Time (s)) & \scriptsize(Energy, Time (s)) & \scriptsize(Energy, Time (s)) & \scriptsize(Energy, Time (s)) \\ 

	\includegraphics[scale = 0.20]{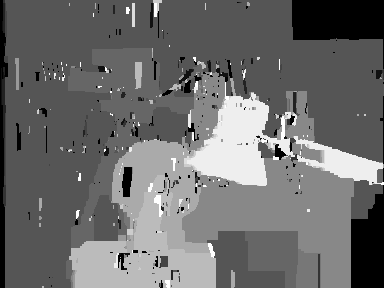} &
	\includegraphics[scale = 0.15]{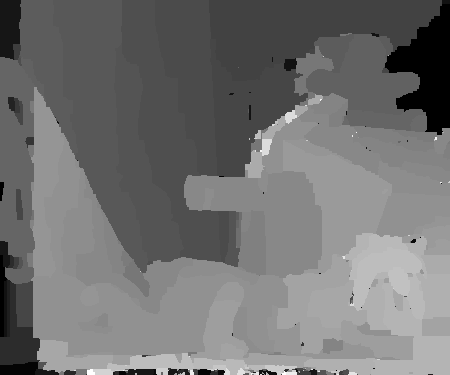} &
	\includegraphics[scale = 0.15]{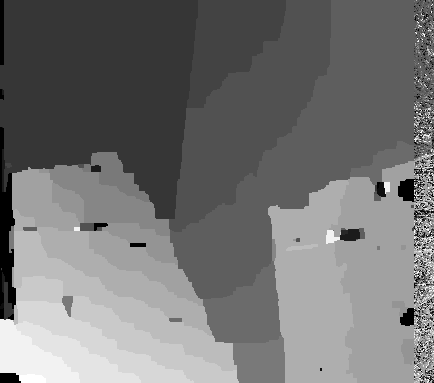} &
	\includegraphics[scale = 0.15]{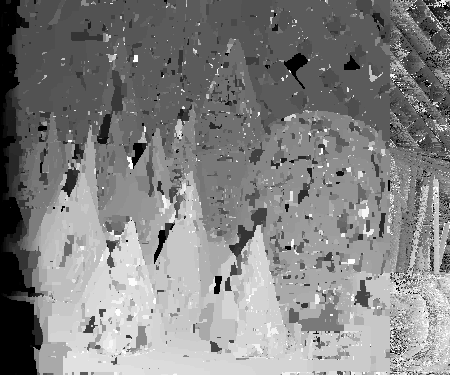} \\
	\scriptsize(1b) Cooccurrence & \scriptsize(2b) Cooccurrence & \scriptsize(3b) Cooccurrence & \scriptsize(4b) Cooccurrence \\
	\scriptsize(2098800, 101) & \scriptsize(3259900, 495) & \scriptsize(2343200, 261) &  \scriptsize(8260100, 308) \\

	\includegraphics[scale = 0.20]{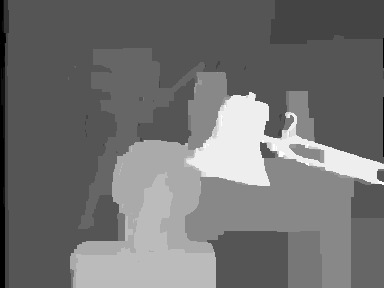} &
	\includegraphics[scale = 0.15]{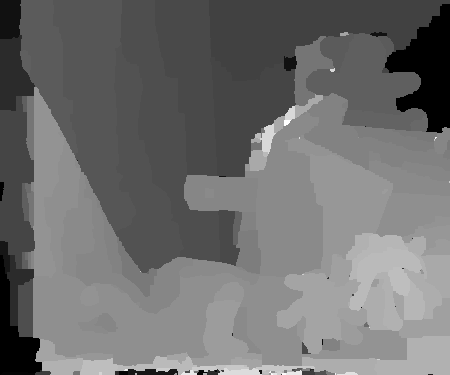} &
	\includegraphics[scale = 0.15]{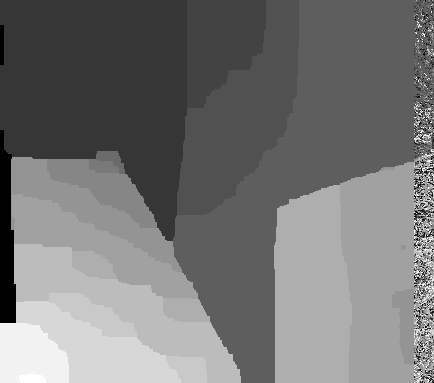} &
	\includegraphics[scale = 0.15]{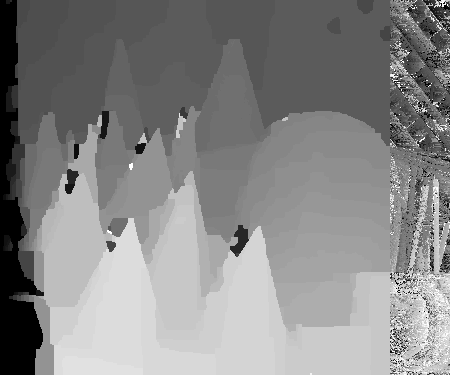} \\
	\scriptsize(1c) Parsimonious & \scriptsize(2c) Parsimonious & \scriptsize(3c) Parsimonious & \scriptsize(4c) Parsimonious \\ 
	\scriptsize(1364200, 225) &  \scriptsize(3201300, 484) & \scriptsize(2262600, 482) & \scriptsize({\bf 4985639}, 759) \\ 

	\includegraphics[scale = 0.20]{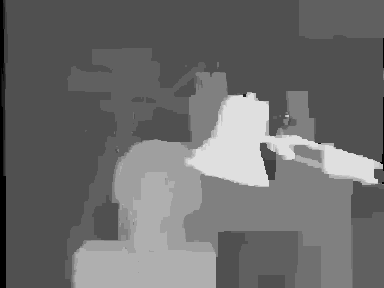} &
	\includegraphics[scale = 0.15]{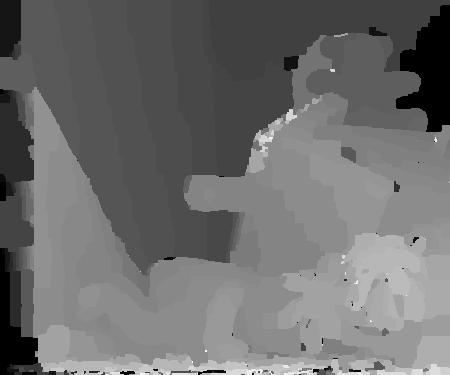} &
	\includegraphics[scale = 0.15]{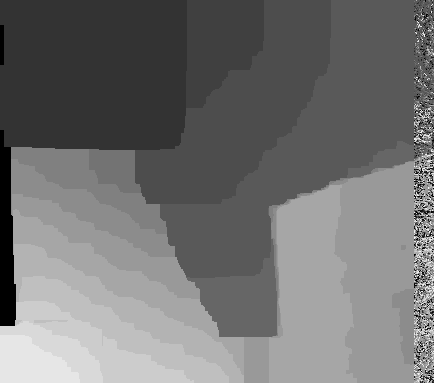} &
	\includegraphics[scale = 0.15]{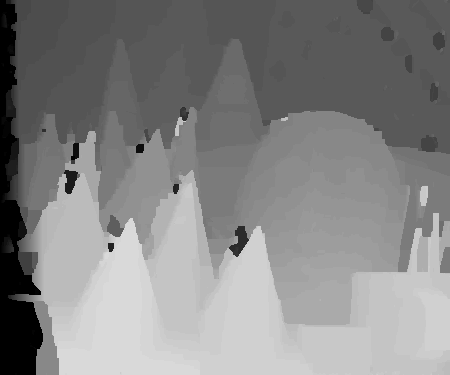} \\
	\scriptsize(1d) $m$ = 1, $h'$ = 4 &  \scriptsize(2d) $m$ = 1, $h'$ = 2 & \scriptsize(3d) $m$ = 1, $h'$ = 4 & \scriptsize(4d) $m$ = 1, $h'$ = 6 \\
	\scriptsize({\bf 1257249}, 256) &  \scriptsize({\bf 3004059}, 1304) & \scriptsize(2210629, 2700) & \scriptsize(5237919, 3711) \\ 

	\includegraphics[scale = 0.20]{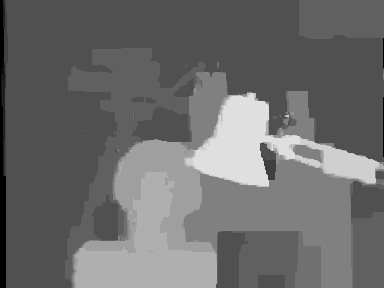} &
	\includegraphics[scale = 0.15]{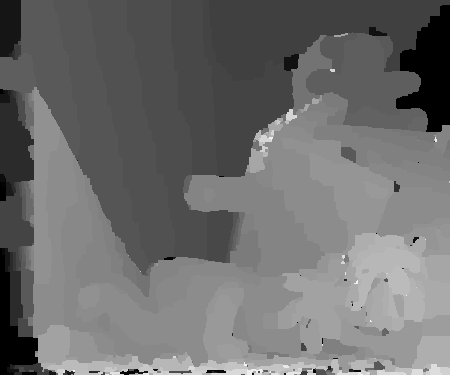} &
	\includegraphics[scale = 0.15]{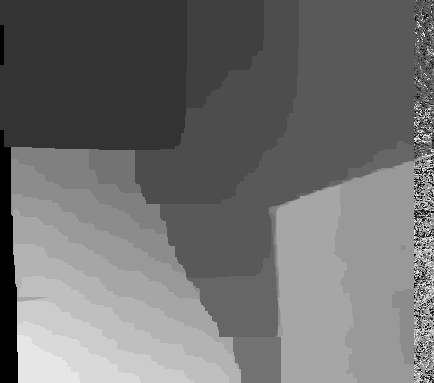} &
	\includegraphics[scale = 0.15]{\imagePath/stereo_results/confFileStereo_coneL6_m1_M5_wc20_labeling} \\
	\scriptsize(1e) $m$ = 1, $h'$ = 6 & \scriptsize(2e) $m$ = 1, $h'$ = 3 & \scriptsize(3e) $m$ = 1, $h'$ = 6 & \scriptsize(4e) $m$ = 1, $h'$ = 6 \\ 
	\scriptsize(1258499, 518) & \scriptsize(3202489, 6216) & \scriptsize({\bf 2207549}, 4624) & \scriptsize(5237919, 3711) \\

	\includegraphics[scale = 0.20]{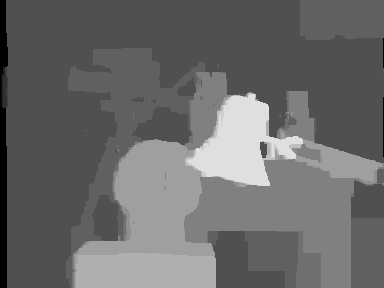} &
	\includegraphics[scale = 0.15]{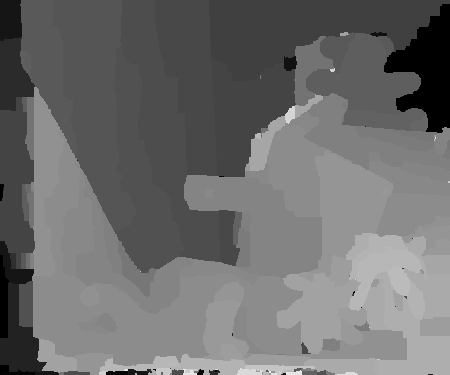} & 
	\includegraphics[scale = 0.15]{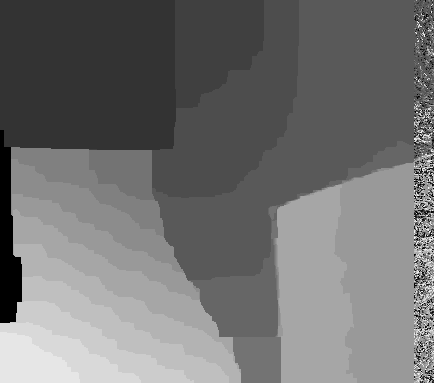} & 
	\includegraphics[scale = 0.15]{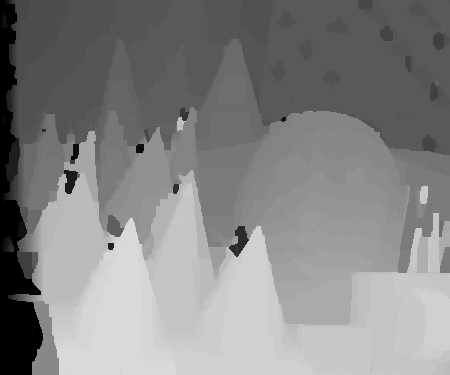} \\ 
	\scriptsize(1f) $m$ = 3, $h'$ = 6 & \scriptsize(2f) $m$ = 3, $h'$ = 3 & \scriptsize(3f) $m$ = 3, $h'$ = 6 & \scriptsize(4f) $m$ = 3, $h'$ = 6 \\
	\scriptsize($1391629^*$, 720) &  \scriptsize($3211819^*$, 5043) &  \scriptsize($2222299^*$, 4189) & \scriptsize($5258999^*$, 4137) \\

	\includegraphics[scale = 0.20]{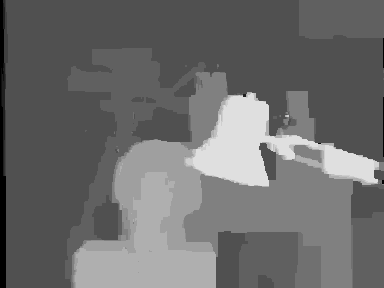} &
	\includegraphics[scale = 0.15]{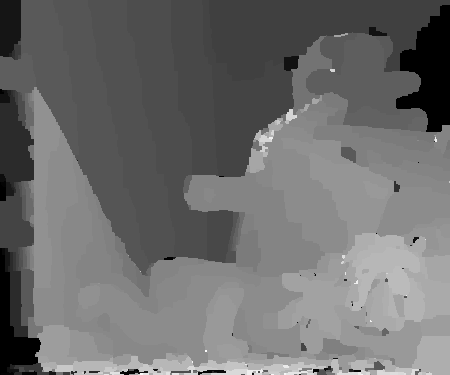} & 
	\includegraphics[scale = 0.15]{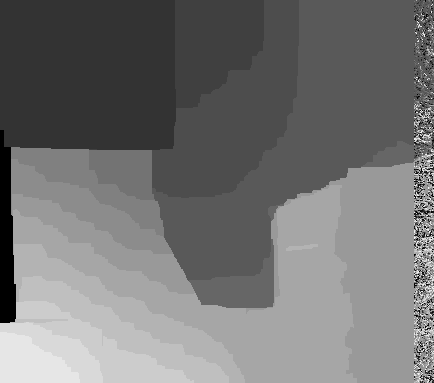} & 
	\includegraphics[scale = 0.15]{\imagePath/stereo_results/confFileStereo_coneL6_m3_M5_wc20_labeling} \\
	\scriptsize(1g) $m$ = 3, $h'$ = 4 &  \scriptsize(2g) $m$ = 3, $h'$ = 2 & \scriptsize(3g) $m$ = 3, $h'$ = 4 & \scriptsize(4g) $m$ = 3, $h'$ = 6 \\
	\scriptsize($1267449^*$, 335) & \scriptsize($3211829^*$, 1139) & \scriptsize($2235689^*$, 3032) & \scriptsize($5258999^*$, 4137) \\

\end{tabular}
\mycaption{\footnotesize \em Stereo matching results: Figures (1a), (1b), (1c) and (1d) are the ground truth disparity for `tsukuba', `teddy', `venus' and `cones' respectively. Our results for $m$ = 1 (1d), (2d), and (3d) are significantly better than those of~\cite{ladickyeccv10} (1b), (2b) and (3b), and of~\cite{dokaniaiccv15} (1c), (2c) and (3c) in terms of energy. For `cones', ~\cite{dokaniaiccv15} (4c) performs better than our algorithm. We also show results for $m$ = 3. We use super-pixels obtained using mean-shift as cliques. $^*$Note that $m$ = 3 uses a different energy function from other cases.}
\label{fig:stereo_matching}
\end{figure}

\myparagraph{\bf Results.} Results for $\omega_c$ = 20, $M$ = 5, and $m$ = 1 and 3 for `tsukuba', `venus' and `cones' and $\omega_c$ = 20, $M$ = 1, and $m$ = 1 and 3 for `teddy' are shown in Figure~\ref{fig:stereo_matching}. We used interval length $h'$ as 4 for `tsukuba' and `venus', 6 for `cones' and 2 for `teddy'. Apart from `cones', our algorithm consistently gives lower energy labeling as compared to both~\cite{dokaniaiccv15} and~\cite{ladickyeccv10}. For `tsukuba', our algorithm captures the details of the face better than~\cite{dokaniaiccv15} and~\cite{ladickyeccv10}. For `venus', we get smoother labeling for the front plane. Moreover, our results for $m$ = 3 exhibit robustness to inaccurate clique definitions.

\myparagraph{\bf Results.} Results for $\omega_c$ = 20, $M$ = 5, and $m$ = 1 and 3 for `tsukuba', `venus' and `cones', and $\omega_c$ = 20, $M$ = 1, and $m$ = 1 and 3 for `teddy' are shown in Figure~\ref{fig:stereo_matching}. We show the results for interval length 3 for `teddy', and 6 for others. Note that in our main paper, we used interval length $h'$ as 4 for `tsukuba' and `venus', and 1 for `teddy' and did not show the results for `cones'. Apart from `cones', our algorithm consistently gives lower energy labeling as compared to both~\cite{dokaniaiccv15} and~\cite{ladickyeccv10}. 

\newgeometry{left=1cm,right=1cm,bottom=1.5cm,top=1.5cm}

\clearpage
\restoregeometry

\mysection{Discussion}
We proposed a novel family of high-order random fields called truncated max-of-convex models (TMCM) which are generalization of truncated convex models (TCM). To perform inference in TMCM, we developed a novel range expansion algorithm for energy minimization that retains the efficiency of $st$-{\sc mincut} and provides provably accurate solutions. The algorithm relies on an exact graph representation of max-of-convex models, a tight submodular overestimate of the energy function for any interval length and a graph construction that represents this overestimate, allowing the inference problem to be solved using $st$-{\sc mincut}. From a theoretical point of view, our work can be thought of as a step towards the identification of graph representable submodular functions and automated construction of graphs for such functions.

{\small
\bibliographystyle{plain}
\bibliography{tmcm}
}

\newpage
\begin{appendices}

\mysection{Proof of Proposition \ref{prop:submodOver}}
\label{app:submodOverest}

{\bf Proposition \ref{prop:submodOver}}: $\delta_{a,b}(y_a,y_b)$ is submodular in the sense of label-set encoding used in ~\cite{ishikawapami03}and is an overestimate of the truncated convex distance $\min\{d(x_a-x_b),M\}$.

\paragraph{Proof:} Note that to prove that $\delta_{a,b}(y_a,y_b)$ is submodular, we need to show that the following inequality holds:

\begin{equation}
\delta_{a,b}(y_a,y_b) + \delta_{a,b}(y_a + 1,y_b + 1) \leq  \delta_{a,b}(y_a + 1,y_b) + \delta_{a,b}(y_a,y_b + 1)
\label{eq:submodInequal}
\end{equation}

We consider the 4 cases corresponding to equation \ref{eq:submodOverestimate}.

{\bf Case 1}: {} $y_a = y_b = 0$ 

\begin{align*}
	\delta_{a,b}(y_a,y_b) &= \delta_{a,b}(0,0) = \min\{d(\hat{x}_a-\hat{x}_b),M\} \\
	\min\{d(x_a-x_b),M\} &= \min\{d(\hat{x}_a-\hat{x}_b),M\} \\
\end{align*}
Hence, $\delta_{a,b}(y_a,y_b) = \min\{d(x_a-x_b),M\}$.

To prove submodularity, consider L.H.S of inequality \ref{eq:submodInequal}:

\begin{align*}
	\delta_{a,b}(y_a,y_b) + \delta_{a,b}(y_a + 1,y_b + 1) &= \delta_{a,b}(0, 0) + \delta_{a,b}(1 , 1) \nonumber \\ 
							&= \min\{d(\hat{x}_a-\hat{x}_b),M\} + 0 
\end{align*}

The R.H.S of inequality \ref{eq:submodInequal} is given by:

\begin{align*}
	\delta_{a,b}(y_a + 1,y_b) + \delta_{a,b}(y_a,y_b + 1) &= \delta_{a,b}(1, 0) + \delta_{a,b}(0 , 1) \nonumber \\
							&= M+d(y_b-1) + M+d(y_a-1) \\
\end{align*}

Clearly L.H.S $<$ R.H.S for this case. 

\bigskip

{\bf Case 2}: {} $y_a = 0, y_b \neq 0$

\begin{align*}
	\delta_{a,b}(y_a,y_b) &= \delta_{a,b}(0,y_b) = M+ d(y_b-1)\\
	\min\{d(x_a-x_b),M\} &= \min\{d(\hat{x}_a-x_b),M\} \\
\end{align*}
Hence, $\delta_{a,b}(y_a,y_b) \geq \min\{d(x_a-x_b),M\}$.

For submodularity, L.H.S of inequality \ref{eq:submodInequal} is given by:

\begin{align*}
	\delta_{a,b}(y_a,y_b) + \delta_{a,b}(y_a + 1,y_b + 1) &= \delta_{a,b}(0, y_b) + \delta_{a,b}(1 , y_b + 1) \nonumber \\ 
							  &= M + d(y_b - 1) + d(y_b)
\end{align*}

The R.H.S of inequality \ref{eq:submodInequal} can be written as:

\begin{align*}
	\delta_{a,b}(y_a + 1,y_b) + \delta_{a,b}(y_a,y_b + 1) &= \delta_{a,b}(1, y_b) + \delta_{a,b}(0 , y_b + 1) \nonumber \\
							&= d(y_b-1) + M+d(y_b) \\
\end{align*}

Hence L.H.S $=$ R.H.S for this case.

{\bf Case 3}: {} $y_a \neq 0, y_b = 0$

\begin{align*}
	\delta_{a,b}(y_a,y_b) &= \delta_{a,b}(y_a,0) = M + d(y_a - 1) \\
	\min\{d(x_a-x_b),M\} &= \min\{d(x_a-\hat{x}_b),M\} \\
\end{align*}
Hence, $\delta_{a,b}(y_a,y_b) \geq \min\{d(x_a-x_b),M\}$.

To prove submodularity, consider L.H.S of inequality \ref{eq:submodInequal}:

\begin{align*}
	\delta_{a,b}(y_a,y_b) + \delta_{a,b}(y_a + 1,y_b + 1) &= \delta_{a,b}(y_a, 0) + \delta_{a,b}(y_a + 1 , 1) \nonumber \\ 
							&= M +  d(y_a - 1) + d(y_a)
\end{align*}

The R.H.S of inequality \ref{eq:submodInequal} is given by:

\begin{align*}
	\delta_{a,b}(y_a + 1,y_b) + \delta_{a,b}(y_a,y_b + 1) &= \delta_{a,b}(y_a + 1, 0) + \delta_{a,b}(y_a , 1) \nonumber \\
							&=  M +  d(y_a) + d(y_a - 1)\\
\end{align*}

Hence L.H.S $=$ R.H.S for this case. 

{\bf Case 4}: {} $y_a \neq 0, y_b \neq 0$

\begin{align*}
	\delta_{a,b}(y_a,y_b) &= d(y_a-y_b)
\end{align*}

Hence, $\delta_{a,b}(y_a,y_b) \geq \min\{d(x_a-x_b),M\}$.

To prove submodularity, consider L.H.S of inequality \ref{eq:submodInequal}:

\begin{align*}
	\delta_{a,b}(y_a,y_b) + \delta_{a,b}(y_a + 1,y_b + 1) = 2 \cdot d(y_a-y_b) 
\end{align*}

The R.H.S of inequality \ref{eq:submodInequal} is given by:

\begin{align*}
	\delta_{a,b}(y_a + 1,y_b) + \delta_{a,b}(y_a,y_b + 1) &= d(y_a-y_b - 1) + d(y_a-y_b + 1)\\
\end{align*}

Since $d()$ is convex, L.H.S $\leq$ R.H.S for this case.

This completes our proof of proposition~\ref{prop:submodOver}. $\blacksquare$
\bigskip

\newpage
\mysection{Proof of Lemma \ref{lemma:cliqueGraphProof}}
\label{app:graphProp}

{\bf Lemma \ref{lemma:cliqueGraphProof}}: Graph (a) of figure~\ref{fig:clique} exactly models clique potentials that are proportional to the sum of $m$ maximum convex distance functions over all disjoint pairs of random variables of the clique.

\paragraph{Proof:} Let us arrange the labels assigned to $\bf{X_c}$ in increasing order as $\{l_{k_{1}}, \dots ,l_{k_{c}}\}$, where $c = |\bf{X_c}|$ indicates the size of the clique. The sum of $m$ maximum convex distance functions over all disjoint pairs of random variables of the clique is given by:

\begin{equation}
\omega_{\bf c}(d(k_{c} - k_{1}) + d(k_{c-1} - k_{2}) + .... + d(k_{c-m+1} - k_{m}))
\end{equation}

When $i \in [k_{p} + 1, k_{p + 1}]$, exactly $p$ variables in ${\bf_{X_c}}$ are assigned a label between [$l_{k_{1}}$, $l_{k_{p}}$] and the corresponding $p$ vertices among $\{V^1_i, V^2_i....V^c_i\}$ will be in $\bf{V}_{t}$ (the remaining $c-t$ vertices will be in $\bf{V}_{s}$). Similarly, when $j \in [k_{c-p} + 1, k_{c-p+1}$], exactly $p$ variables in ${\bf_{X_c}}$ are assigned a label in the range [$l_{k_{c - p + 1}}$, $l_{k_{c}}$] and the corresponding $p$ vertices among $\{V^1_j, V^2_j....V^c_j\}$ will be in $\bf{V}_{s}$ (the remaining $c-p$ vertices will be in $\bf{V}_{t}$). $st$-{\sc mincut} includes either the arcs $(U_{ij}, V_{i}^{a})$ for all $V_{i}^{a} \in \bf{V}_{t}$ or the arcs $(V_{j}^{a}, W_{ij})$ for all $V_{j}^{a} \in \bf{V}_{s}$ or the arc $(W_{ij}, U_{ij})$, whichever contribute the smallest cost to the cut. Let $C_{ij}$ denote the sum of capacitites of arcs cut by $st$-{\sc mincut} in graph (a) of figure~\ref{fig:clique} for a given pair $i, j$. Clearly, $ 0 \leq C_{ij} \leq m \cdot r_{ij}$.

Let us partition the interval $[k_{1}+1, k_{c}]$ to $2m-1$ intervals $[k_{1}+1, k_{2}], [k_{2}+1, k_{3}],...[k_{m}+1, k_{c-m+1}]...,[k_{c-2}+1, k_{c-1}], [k_{c-1}+1, k_{c}] $. $C_{ij}$ depends on the intervals in which $i$ and $j$ lie. Given a labeling $\bf{x_{c}}$, the total contribution of all the cut arcs is equal to

\begin{align}
	\sum_{i=k_{1}+1}^{k_{c}} \sum_{j=i}^{k_{c}} C_{ij} &= \sum_{i=k_{1}+1}^{k_{2}} \sum_{j=i}^{k_{c}} C_{ij} +
\sum_{i=k_{2}+1}^{k_3} \sum_{j=i}^{k_{c}} C_{ij} \nonumber \\ &+\dotsb+  
\sum_{i=k_{m}+1}^{k_{c-m+1}} \sum_{j=i}^{k_{c}} C_{ij} +\dotsb+ \sum_{i=k_{c-1}+1}^{k_{c}} \sum_{j=i}^{k_{c}} C_{ij}
\label{eq:partition}
\end{align}

Consider the second term of the series in (\ref{eq:partition}). When $i \in [k_2 + 1, k_3]$, exactly 2 variables in ${\bf{X_c}}$ take up a label in the range $[k_1, k_2]$ and hence, two vertices among $\{V^1_i, V^2_i....V^c_i\}$ belong to $\bf{V}_{t}$. When $j \in [k_{c-1} + 1, k_c]$, only one vertex among $\{V^1_j, V^2_j....V^c_j\}$ belongs to $\bf{V}_{s}$, giving $C_{ij}$ as $r_{ij}$. For $j$ in all other intervals, at least two vertices among $\{V^1_j, V^2_j....V^c_j\}$ belong to $\bf{V}_{s}$, and $C_{ij}$ equals $2 \cdot r_{ij}$. Thus, the following holds

\begin{align*}
\sum_{i=k_{2}+1}^{k_3} \sum_{j=i}^{k_{c}} C_{ij} &= \sum_{i=k_{2}+1}^{k_3} \sum_{j=k_{c-1} + 1}^{k_{c}} r_{ij} + \sum_{i=k_{2}+1}^{k_3} \sum_{j=i}^{k_{c-1}} 2r_{ij} \\
&= \sum_{i=k_{2}+1}^{k_3} \sum_{j=k_{c-1} + 1}^{k_{c}} r_{ij} + \left(\sum_{i=k_{2}+1}^{k_3} \sum_{j=i}^{k_{c-1}} r_{ij} + \sum_{i=k_{2}+1}^{k_3} \sum_{j=i}^{k_{c-1}} r_{ij}\right)\\
&= \sum_{i=k_{2}+1}^{k_3} \sum_{j=i}^{k_{c}} r_{ij} + \sum_{i=k_{2}+1}^{k_3} \sum_{j=i}^{k_{c-1}} r_{ij}
\end{align*} 

where the last step is obtained by combining the first and the third terms in the summation.

In general, for $p \in [1, m-1]$ we can write

\begin{equation}
\sum_{i=k_{p}+1}^{k_{p+1}} \sum_{j=i}^{k_{c}} C_{ij} = \sum_{i=k_{p}+1}^{k_{p+1}} \sum_{j=i}^{k_{c}} r_{ij} + \sum_{i=k_{p}+1}^{k_{p+1}} \sum_{j=i}^{k_{c-1}} r_{ij} +\dotsb+ \sum_{i=k_{p}+1}^{k_{p+1}} \sum_{j=i}^{k_{c-p+1}} r_{ij}
\label{eq:regrouping}
\end{equation}

Similar argument can be extended for $p \in [c-m+1, c-1]$
 
Using (\ref{eq:regrouping}), each term of (\ref{eq:partition}) can be written as

\begin{align*}
\sum_{i=k_{1}+1}^{k_{2}} \sum_{j=i}^{k_{c}} C_{ij} &= \sum_{i=k_{1}+1}^{k_{2}} \sum_{j=i}^{k_{c}} r_{ij}\\
\sum_{i=k_{2}+1}^{k_{3}} \sum_{j=i}^{k_{c}} C_{ij} &= \sum_{i=k_{2}+1}^{k_{3}} \sum_{j=i}^{k_{c}} r_{ij} + \sum_{i=k_{2}+1}^{k_{3}} \sum_{j=i}^{k_{c-1}} r_{ij}\\
\sum_{i=k_{3}+1}^{k_{4}} \sum_{j=i}^{k_{c}} C_{ij} &= \sum_{i=k_{3}+1}^{k_{4}} \sum_{j=i}^{k_{c}} r_{ij} + \sum_{i=k_{3}+1}^{k_{4}} \sum_{j=i}^{k_{c-1}} r_{ij} + \sum_{i=k_{3}+1}^{k_{4}} \sum_{j=i}^{k_{c-2}} r_{ij}\\
\vdots\\
\sum_{i=k_{m}+1}^{k_{c-m+1}} \sum_{j=i}^{k_{c}} C_{ij} &= \sum_{i=k_{m}+1}^{k_{c-m+1}} \sum_{j=i}^{k_{c}} r_{ij} + \sum_{i=k_{m}+1}^{k_{c-m+1}} \sum_{j=i}^{k_{c-1}} r_{ij} +\dotsb\dotsb+ \sum_{i=k_{m}+1}^{k_{c-m+1}} \sum_{j=i}^{k_{c-m+1}} r_{ij}\\
\vdots\\
\sum_{i=k_{c-3}+1}^{k_{c-2}} \sum_{j=i}^{k_{c}} C_{ij} &= \sum_{i=k_{c-3}+1}^{k_{c-2}} \sum_{j=i}^{k_{c}} r_{ij} + \sum_{i=k_{c-3}+1}^{k_{c-2}} \sum_{j=i}^{k_{c-1}} r_{ij} + \sum_{i=k_{c-3}+1}^{k_{c-2}} \sum_{j=i}^{k_{c-2}} r_{ij}\\
\sum_{i=k_{c-2}+1}^{k_{c-1}} \sum_{j=i}^{k_{c}} C_{ij} &= \sum_{i=k_{c-2}+1}^{k_{c-1}} \sum_{j=i}^{k_{c}} r_{ij} + \sum_{i=k_{c-2}+1}^{k_{c-1}} \sum_{j=i}^{k_{c-1}} r_{ij}\\
\sum_{i=k_{c-1}+1}^{k_{c}} \sum_{j=i}^{k_{c}} C_{ij} &= \sum_{i=k_{c-1}+1}^{k_{c}} \sum_{j=i}^{k_{c}} r_{ij}
\end{align*}
Adding all the above equations by adding all terms vertically, we obtain
\begin{align*}
\sum_{i=k_{1}+1}^{k_{c}} \sum_{j=i}^{k_{c}} C_{ij} &= \sum_{i=k_{1}+1}^{k_{c}} \sum_{j=i}^{k_{c}} r_{ij} +
\sum_{i=k_{2}+1}^{k_{c-1}} \sum_{j=i}^{k_{c-1}} r_{ij} +\dotsb+ 
\sum_{i=k_{m}+1}^{k_{c-m+1}} \sum_{j=i}^{k_{c-m+1}} r_{ij}\\\\
&= \omega_{\bf{c}}d(k_{c}-k_{1}) + \omega_{\bf{c}}d(k_{c-1}-k_{2}) +\dotsb+
\omega_{\bf{c}}d(k_{c-m+1}-k_{m})
\end{align*}

Hence, graph (a) of figure~\ref{fig:clique} represents exactly clique potentials that are proportional to the sum of $m$ maximum convex distance functions over all disjoint pairs of random variables of the clique. \hspace{6mm} $\blacksquare$

\newpage
\mysection{Proof of Lemma \ref{lemma:reductionlowerBound}}
\label{app:lemmaProofreduction}
{\bf Lemma \ref{lemma:reductionlowerBound}:} At an iteration of our algorithm, given the current labeling $f_n$ and an interval $I_n = [i_n + 1, j_n]$, the new labeling $f_{n+1}$ obtained by solving the st-mincut problem reduces the energy by at least the following:

\begin{equation*}
	\begin{split}
		& \sum_{X_a \in {\bf X}(f^*, I_{n})} \theta_{a}(f_n(a)) + \sum_{{\bf X_c} \in \mathcal{A}\mathit(f, I_n) \cup \mathcal{B}\mathit(f, I_n)} \theta_{\bf c}({\bf x_c}) \\
		&	  - \left( \sum_{X_a \in {\bf X}(f^*, I_{n})} \theta_{a}(f^*(a)) + \sum_{{\bf X_c} \in \mathcal{A}\mathit(f, I_n)} t_c^n 
	 +  \sum_{{\bf X_c} \in \mathcal{B}\mathit(f, I_n)} s_c^n \right) 
	 \end{split}
\end{equation*}

\paragraph{Proof:} It is evident from the properties of the graph discussed in subsection~\ref{subsec:graph_properties} that the energy of the new labeling $f_{n+1}$ is bounded from above by the cost of the $st$-mincut. The cost of the $st$-mincut itself is bounded from above by the cost of any other $st$-cut in the graph. Consider one such $st$-cut that gives the following labeling:

\begin{align*}
 f(a) = 
	\begin{cases}
		f^*(a) & if \hspace{2mm} X_a \in \mathbf{X}(f^*, I_n) \\
		f_n(a) & otherwise.
	\end{cases}
\end{align*}

We can derive the cost of this $st$-cut using the properties of subsection~\ref{subsec:graph_properties}. The energy of $f$ equals the sum of the terms in the properties minus $\sum_{c \in \mathcal{C}} \kappa_{c}$. The energy of $f_{n+1}$ is less than that of $f$. Hence, the difference between the energy of the current labeling $f_n$ and the new labeling $f_{n+1}$, i.e $E(f_n) - E(f_{n+1})$, is at least

\begin{align*}
	\begin{split}
		& \sum_{X_a \in {\bf X}(f^*, I_{n})} \theta_{a}(f_n(a)) + \sum_{{\bf X_c} \in \mathcal{A}\mathit(f, I_n) \cup \mathcal{B}\mathit(f, I_n)} \theta_{\bf c}({\bf x_c}) \\
		&	  - \left( \sum_{X_a \in {\bf X}(f^*, I_{n})} \theta_{a}(f^*(a)) + \sum_{{\bf X_c} \in \mathcal{A}\mathit(f, I_n)} t_c^n 
	 +  \sum_{{\bf X_c} \in \mathcal{B}\mathit(f, I_n)} s_c^n \right) 
	 \end{split}
 \end{align*}

 This proves the lemma. \hspace{6mm} $\blacksquare$ 

\newpage
\mysection{Proof of Lemma \ref{lemma:linearIneq}}
\label{app:linearIneqProof}

{\bf Lemma \ref{lemma:linearIneq}:} When $d(.)$ is linear, that is, $d(x) = |x|$, the following inequality holds true:
\begin{equation}
\begin{split}
& \frac{1}{L}\sum_{r}\sum_{I_n \in \Gamma_r}\left(\sum_{{\bf X_c} \in \mathcal{A}\mathit(f, I_n)} t_c^n + \sum_{{\bf X_c} \in \mathcal{B}\mathit(f, I_n)} s_c^n \right) \\
& \leq  max \left\{ \frac{c}{2}\left(2 + \frac{L}{M}\right), \left(2 + \frac{2M}{L}\right) \right\} \sum_{{\bf c} \in \mathcal{C}} \theta_{\bf c}(\bf {x_c})
\end{split}
\end{equation}
where $\mathit{c}$ is the largest clique in the random field.

\paragraph{Proof:} Let us denote the set of optimum labels in a clique $\mathbf{X_{c}}$ arranged in an increasing order as ${l_1, l_2....., l_{c-1}, l_c}$. For notational simplicity, we introduce  a new notation $L = h'$. Since we are dealing with the truncated linear metric, the terms $t_c^n$ and $s_c^n$ can be simplified as:

\begin{equation}
	t_c^n = \omega_c (l_c - l_1), s_c^n = \omega_c (l_c - i_n - 1 + M)
\end{equation}

The LHS of inequality in lemma \ref{lemma:linearIneq} can be written as:

\begin{equation}
\frac{1}{L}\sum_{{\bf c} \in \mathcal{C}}\left(\sum_{\mathcal{A}\mathit(f, I_n) \ni {\bf c}} t_c^n + \sum_{\mathcal{B}\mathit(f, I_n) \ni {\bf c}} s_c^n \right)
\label{eq:rewriteIneq}
\end{equation}

In order to prove the lemma, we consider the following three cases for each clique ${\bf X_c}$: \\

$Case - I: |l_c - l_1| \geq L$ and hence $\theta_{\bf c}(\bf{x_c}) = \omega_c \cdot M$

In this case, ${\bf X_c} \notin \mathcal{A}\mathit(f, I_n)$ for all intervals $I_n$ since the length of each interval is $L$. Moreover, the conditions for ${\bf X_c} \in \mathcal{B}\mathit(f, I_n)$ are given by

\begin{equation*}
	{\bf X_c} \in \mathcal{B}\mathit(f, I_n) \iff i_n \in [l_1 - L, l_c - 1]
\end{equation*}
We observe that

\begin{align}
	& \sum_{\mathcal{A}\mathit(f, I_n) \ni {\bf c}} t_c^n + \sum_{\mathcal{B}\mathit(f, I_n) \ni {\bf c}} s_c^n \nonumber\\
	\leq{}  & \omega_c \sum_{p = 1}^{c} \sum_{l_p - L}^{l_p - 1} \left( M + l_p - i_n - 1 \right)\nonumber\\        
	={}  & \omega_c \sum_{p=1}^{c} \left(L \cdot M + \frac{L \cdot (L - 1)}{2}\right) \nonumber\\
	={}  & L \cdot \frac{c}{2} \left( 2 + \frac{L - 1}{M} \right) \theta_{\bf c}(\bf{x_c}) \nonumber\\
	\leq{}  & L \cdot \frac{c}{2} \left( 2 + \frac{L}{M} \right) \theta_{\bf c}(\bf{x_c}) \label{eq:case1bound}
\end{align}
	
$Case - II: M \leq |l_c - l_1| < L$ and hence $\theta_{\bf c}(\bf{x_c}) = \omega_c \cdot M$

In this case, the conditions for ${\bf X_c} \in \mathcal{A}\mathit(f, I_n)$ and $X_c \in \mathcal{B}\mathit(f, I_n)$ are given by

\begin{align*}
& {\bf X_c} \in \mathcal{A}\mathit(f, I_n) \iff i_n \in [l_c - L, l_1 - 1] \\
& {\bf X_c} \in \mathcal{B}\mathit(f, I_n) \iff i_n \in [l_1 - L, l_c - L - 1] \cup [l_1, l_c - 1]\\
\end{align*}
We observe that

\begin{align*}
	& \sum_{\mathcal{A}\mathit(f, I_n) \ni {\bf c}} t_c^n + \sum_{\mathcal{B}\mathit(f, I_n) \ni {\bf c}} s_c^n \\
	={}  & \omega_c \left( \sum_{i_n = l_c - L}^{l_1 - 1} (l_c - l_1) + \left\{ \sum_{p = 1}^{c-1} \sum_{i_n = l_p - L}^{l_{p+1}-L-1} (M + l_p - (i_n + 1)) \right.\right.\\ 
	     & \left. \left. + \sum_{i_n = l_1}^{l_c -1} (M + l_c - (i_n + 1)) \right\}\right)\\        
	={} & \omega_c \left( A + B + C \right) \\
\end{align*}
where
\begin{align*}
	A  ={} &\sum_{i_n = l_c - L}^{l_1 - 1} (l_c - l_1) \\
	B  ={} &\sum_{p = 1}^{c-1} \sum_{i_n = l_p - L}^{l_{(p+1)}-L-1} (M + l_p - (i_n + 1)) \\
	C  ={} &\sum_{i_n = l_1}^{l_c -1} (M + l_c - (i_n + 1)) \\
\end{align*}
Let $(l_c - l_1) = r$. Hence, we have:

\begin{align}
	A & = (l_c - l_1)\cdot(L - (l_c - l_1)) \nonumber\\
	  & = r \cdot (L - r) = L \cdot r - r^2 \label{eq:Aexpr}
\end{align}

\begin{align*}
	& B = \sum_{p = 1}^{c-1} \sum_{i_n = l_p - L}^{l_{(p+1)}-L-1} (M + l_p - (i_n + 1)) \\
	&  = \sum_{p = 1}^{c-1} \sum_{i_n = l_p - L}^{l_{(p+1)}-L-1} M + \\
	& \sum_{p = 1}^{c-1} \sum_{i_n = l_p - L}^{l_{(p+1)}-L-1} (l_p - (i_n + 1)) \\
	& = \sum_{p = 1}^{c-1} M \cdot (l_{(p+1)} - l_p) +  \\
	& \sum_{p = 1}^{c-1}\left\{(L-1) +.....+ (L-(l_{p+1}-l_p))\right\} \\
\end{align*}

Let $ y_p = l_{(p+1)} - l_p $. Clearly 
\begin{equation*}
\sum_{p=1}^{c-1} y_p = l_c - l_1 \\
\end{equation*}	

Hence,

\begin{align}
 B & = M \cdot r + L \cdot r - \sum_{p=1}^{c-1} \frac{y_p \cdot (y_p + 1)}{2} \nonumber \\
	  & = M \cdot r + L \cdot r - \frac{r}{2} - \sum_{p=1}^{c-1} \frac{(y_p)^2}{2} \label{eq:Bexpr}
\end{align}

\begin{align}
	C & = \sum_{i_n = l_1}^{l_c - 1} (M + l_c - (i_n + 1)) \nonumber\\
	  & = \sum_{i_n = l_1}^{l_c - 1} M + \sum_{i_n = l_1}^{l_c - 1} \left(l_c - (i_n + 1) \right) \nonumber\\
	  & = M\cdot(l_c - l_1) + \left\{ (l_c - (l_1 + 1)) + (l_c - (l_1 + 2))... + 1 \right\} \nonumber \\
	& = M\cdot(l_c - l_1) + \frac{(l_c - l_1 - 1)(l_c - l_1)}{2} \nonumber \\
	& = M \cdot r + \frac{(r-1)\cdot r}{2} \nonumber \\
	& = M \cdot r + \frac{r^2}{2} - \frac{r}{2} \label{eq:Cexpr}
\end{align}

Using \eqref{eq:Aexpr}, \eqref{eq:Bexpr}, \eqref{eq:Cexpr}, we have

\begin {equation} \label{eq:sum_ABC}
A + B + C = (2L + 2M - r)r + \frac{r^2}{2} - r -\sum_{p=1}^{c-1}\frac{y_{p}^2}{2}
\end{equation}

We have
\begin{equation*}
	\sum_{p = 1}^{c-1} \frac{{y_p}^2}{2} \geq \frac{r^2}{2(c-1)} \\
\end{equation*}
(Taking each $y_p = r/(c-1)$)


Hence,

\begin{equation*}
	A + B + C \leq \left\{2L + 2M - 1 - \frac{r}{2}\left(\frac{c}{c - 1}\right)\right\}\cdot r
\end{equation*}

Let 

\begin{equation*}
	f(r) = \left\{2L + 2M - 1 - \frac{r}{2} \left(\frac{c}{c - 1}\right)\right\}\cdot r
\end{equation*}

Taking derivative of $f(r)$, we obtain

\begin{equation*}
	f'(r) = 2L + 2M - 1 - \frac{r \cdot c}{(c-1)} 
\end{equation*}

$f(r)$ is a quadratic function which opens downwards (coefficient of $r^2$ is negative). Also

\begin{equation*}
	f'(L) = \left(\frac{c - 2}{c - 1}\right) \cdot L + 2M - 1 > 0 
\end{equation*}

Hence, using $r = L$,

\begin{align}
A + B + C & < L \cdot \left\{2L + 2M - 1 - \frac{L}{2} \cdot \left( \frac{c}{c - 1} \right) \right\} \nonumber\\
			       & < L \cdot \left\{\frac{2L}{M} + 2 - \frac{L}{2M} \cdot \left( \frac{c}{c - 1} \right) \right\} \theta_{\bf c}(\bf{x_c}) \label{eq:case2bound}
\end{align}
where the last expression is obtained using the fact that $\theta_{\bf c}(\bf{x_c}) = \omega_c \cdot M$. \\

$Case - III: |l_c - l_1| < M$ and hence $\theta_{\bf c}(\bf{x_c}) = \omega_c(l_c - l_1)$

In this case, the conditions for ${\bf X_c} \in \mathcal{A}\mathit(f, I_n)$ and ${\bf X_c} \in \mathcal{B}\mathit(f, I_n)$ are given by

\begin{align*}
& {\bf X_c} \in \mathcal{A}\mathit(f, I_n) \iff i_n \in [l_c - L, l_1 - 1] \\
& {\bf X_c} \in \mathcal{B}\mathit(f, I_n) \iff i_n \in [l_1 - L, l_c - L - 1] \cup [l_1, l_c - 1]\\
\end{align*}

We observe that

\begin{align*}
	& \sum_{\mathcal{A}\mathit(f, I_n) \ni {\bf c}} t_c^n + \sum_{\mathcal{B}\mathit(f, I_n) \ni {\bf c}} s_c^n \\
	={}  & \omega_c \left( \sum_{i_n = l_c - L}^{l_1 - 1} (l_c - l_1) + \left\{ \sum_{p = 1}^{c-1} \sum_{i_n = l_p - L}^{l_{p+1}-L-1} (M + l_p - (i_n + 1)) \right.\right.\\ 
	     & \left. \left. + \sum_{i_n = l_1}^{l_c -1} (M + l_c - (i_n + 1)) \right\}\right)\\        
	={} & \omega_c \left( A + B + C \right) \\
\end{align*}

Similar to case II, we can write

\begin{equation*} \label{eq:rewrite_sum_ABC}
A + B + C = (2L + 2M - r/2)r - r -\sum_{p=1}^{c-1}\frac{y_{p}^2}{2} \\
\end{equation*}

Since $r < M$

\begin{align}
	A + B + C &\leq (2L + 2M)(l_{c} - l_{1}) \nonumber\\
		  & = L\left(2 + \frac{2M}{L}\right) \theta_{\bf c}(\bf{x_c}) \label{eq:case3bound}
\end{align}
where the last expression is obtained using the fact that $\theta_{\bf c}(\bf{x_c}) = \omega_c \cdot M$.

Substituting inequalities \eqref{eq:case1bound}, \eqref{eq:case2bound} and \eqref{eq:case3bound} in expression \eqref{eq:rewriteIneq} and dividing both sides by $L$ for all $\bf{X_c}$, we obtain inequality of lemma \ref{lemma:linearIneq}. This proves the lemma. \hspace{6mm} $\blacksquare$ 

\newpage
\mysection{Proof of Proposition \ref{prop:linearBound}}
\label{app:linearBoundProof}

{\bf Proposition \ref{prop:linearBound}:} The range expansion algorithm with $h' = M$ has a multiplicative bound of $O(C)$ for truncated max-of-linear model when $m=1$. The
term $C$ equals the size of the largest clique. Hence, if ${\bf x}^*$ is a labeling with
minimum energy and $\hat{\bf x}$ is the labeling estimated by range expansion algorithm then
\begin{equation}
\small
\sum_{a \in {\cal V}} \theta_a(\hat{x}_a) + \sum_{{\bf c} \in {\cal C}} \theta_{\bf c}(\hat{\bf x}_{\bf c}) \leq \nonumber 
\sum_{a \in {\cal V}} \theta_a({x}^*_a) + O(C) \sum_{{\bf c} \in {\cal C}} \theta_{\bf c}({\bf x}^*_{\bf c}).
\end{equation}
The above inequality holds for arbitrary set of unary potentials and non-negative clique weights.

\paragraph{Proof:} The following equation can be deduced from the above definitions:

\begin{equation} \label{unary_sum}
	\sum_{X_a \in \mathbf{X}} \theta_a(f^*(a)) = \sum_{I_n \in \mathcal{I}_r} \sum_{X_a \in \mathbf{X}(f^*, I_{n})} \theta_{a}(f^*(a))
\end{equation}

since $f^*(a)$ belongs to exactly one interval in $I_r$ for all $X_a$. 

For the final labeling $f$ of the range expansion algorithm, the term in lemma~\ref{lemma:reductionlowerBound} should be non-positive for all intervals $I_n$ because $f$ is a local optimum. Hence,

\begin{equation*}
	\begin{split}
		&\sum_{X_a \in {\bf X}(f^*, I_{n})} \theta_{a}(f(a)) + \sum_{{\bf X_c} \in \mathcal{A}\mathit(f, I_n) \cup \mathcal{B}\mathit(f, I_n)} \theta_{\bf c}({\bf x_c})\\
	 & \leq \left( \sum_{X_a \in {\bf X}(f^*, I_{n})} \theta_{a}(f^*(a)) + \sum_{{\bf X_c} \in \mathcal{A}\mathit(f, I_n)} t_c^n 
	 +  \sum_{{\bf X_c} \in \mathcal{B}\mathit(f, I_n)} s_c^n \right), \forall I_{n} 
	\end{split}
\end{equation*}

We sum the above inequality over all $I_n \in \Gamma_r$. The summation of the LHS is at least $E(f)$. Also, using \eqref{unary_sum}, the summation of the above inequality can be written as:

\begin{equation*}
E(f) \leq \sum_{X_a \in {\bf X}} \theta_{a}(f^*(a)) + \sum_{I_n \in \Gamma_r}\left(\sum_{{\bf X_c} \in \mathcal{A}\mathit(f, I_n)} t_c^n + \sum_{{\bf X_c} \in \mathcal{B}\mathit(f, I_n)} s_c^n \right)
\end{equation*}

We now take the expectation of the above inequality over the uniformly distributed random integer $r \in [0, L - 1]$. The LHS of the inequality and the first term on the RHS (that is, $\sum \theta_a(f^*(a)))$ are constants with respect to $r$. Hence, we get

\begin{equation}
E(f) \leq \sum_{X_a \in {\bf X}} \theta_{a}(f^*(a)) + \frac{1}{L}\sum_{r}\sum_{I_n \in \Gamma_r}\left(\sum_{{\bf X_c} \in \mathcal{A}\mathit(f, I_n)} t_c^n + \sum_{{\bf X_c} \in \mathcal{B}\mathit(f, I_n)} s_c^n \right)
\end{equation} 

Lemma~\ref{lemma:linearIneq} allows us to write the above inequality as 

\begin{equation}
E(f) \leq \sum_{X_a \in {\bf X}} \theta_{a}(f^*(a)) + max \left\{ \frac{c}{2}\left(2 + \frac{L}{M}\right), \left(2 + \frac{2M}{L}\right) \right\} \sum_{{\bf c} \in \mathcal{C}} \theta_{\bf c}(\bf {x_c})
\end{equation}

The R.H.S of lemma \ref{lemma:linearIneq} is minimum under the following condition:

\begin{equation}
\label{eq:linearboundEquality}
\frac{c}{2}\left(2 + \frac{L}{M}\right) =  \left(2 + \frac{2M}{L}\right)
\end{equation}
The positive solution of the above quadratic equation gives the optimum value $L_{opt}$ of interval length:

\begin{equation}
	L_{opt} = \left\{\frac{2 - c + \sqrt{c^2 + 4}}{c}\right\}\cdot M
\label{eq:Loptimum}
\end{equation}

However, we bound the minimum value of L as M, giving:

\begin{equation}
	L_{opt} = max\left\{\left\{\frac{2 - c + \sqrt{c^2 + 4}}{c}\right\}\cdot M, M\right\}
\label{eq:Loptimumbound}
\end{equation}

Note $L_{opt}$ equals $\sqrt{2}M$ for $c = 2$ and $M$ for $c > 2$. Substituting the optimum value of L from equation \ref{eq:Loptimumbound} in inequality of lemma \ref{lemma:linearIneq} and simplifying inequality \ref{eq:upperboundfinalenergy} gives the multiplicative bound for truncated max-of-linear model as $\frac{(c+2) + \sqrt{c^2 + 4}}{2}$. Hence, the multiplicative bound is $O(c)$ where $c$ is the size of the maximal clique. For $c = 2$, this gives a bound of 2 + $\sqrt{2}$ using $L = \sqrt{2}M$. \hspace{6mm} $\blacksquare$ 

\vspace{4mm}

\newpage
\mysection{Proof of Proposition \ref{prop:mlinearBound}}
\label{app:mlinearBoundProof}

{\bf Proposition \ref{prop:mlinearBound}}: The range expansion algorithm with $h' = M$ has a multiplicative bound of $O(m \cdot C)$ for truncated max-of-linear model for any general value of $m$. The term $C$ equals the size of the largest clique. Hence, if ${\bf x}^*$ is a labeling with
minimum energy and $\hat{\bf x}$ is the labeling estimated by range expansion algorithm then

\begin{equation}
\small
\sum_{a \in {\cal V}} \theta_a(\hat{x}_a) + \sum_{{\bf c} \in {\cal C}} \theta_{\bf c}(\hat{\bf x}_{\bf c}) \leq \nonumber 
\sum_{a \in {\cal V}} \theta_a({x}^*_a) + O(m \cdot C) \sum_{{\bf c} \in {\cal C}} \theta_{\bf c}({\bf x}^*_{\bf c}).
\label{eq:mlinear_bound}
\end{equation}

The above inequality holds for arbitrary set of unary potentials and non-negative clique weights.

\paragraph{Proof:} We introduce some notations for our proof. Let $\mathcal{A}\mathit(f, I_n)$ be the set of all cliques for which all variables take optimum label in the interval $I_{n}$ and $\mathcal{B}_k\mathit(f, I_n)$ where $k \in [1, \dots, m -1]$ be the set of all cliques for which exactly $k$ variables retain their old label. Let $\mathcal{B}_m\mathit(f, I_n)$ be the set of all cliques for which $m$ or more variables retain their old label.

We also introduce the following shorthand notation:

\begin{itemize}
	\item We denote $\mathbf{\omega_c}\{\sum_{i = 1}^{m} d(p_i({\bf x}_c) - p_{c - i + 1}({\bf x}_c)\} $ as $t_{m, c}^n$
	\item We denote $\mathbf{\omega_c}\{\sum_{i = 2}^{m - 1} d(p_i({\bf x}_c) - p_{c - i + 1}({\bf x}_c) +  d(p_{c}({\bf x}_c) - i_m - 1) \} +  \mathbf{\omega_c}\cdot M$ as $s_{1, c}^n$
	\item In general, we denote $\mathbf{\omega_c}\{\sum_{i = k + 1}^{m - k} d(p_i({\bf x}_c) - p_{c - i + 1}({\bf x}_c) + \sum_{i = 1}^{k} d(p_{c - i + 1}({\bf x}_c) - i_m - 1) \} + $k$ \cdot  \mathbf{\omega_c}\cdot M$ as $s_{k, c}^n$
\end{itemize}

We state the following lemma which is a generalization of lemma~\ref{lemma:reductionlowerBound}.

\begin{lemma}
At an iteration of our algorithm, given the current labeling $f_n$ and an interval $I_n = [i_n + 1, j_n]$, the new labeling $f_{n+1}$ obtained by solving the st-mincut problem reduces the energy by at least the following:

\begin{equation*}
	\begin{split}
		& \sum_{X_a \in {\bf X}(f^*, I_{n})} \theta_{a}(f_n(a)) + \sum_{{\bf X_c} \in \mathcal{A}\mathit(f, I_n) \cup \mathcal{B}_1\mathit(f, I_n) \cup \dots  \cup \mathcal{B}_m\mathit(f, I_n)} \theta_{\bf c}({\bf x_c}) \\
		&	  - \left( \sum_{X_a \in {\bf X}(f^*, I_{n})} \theta_{a}(f^*(a)) + \sum_{{\bf X_c} \in \mathcal{A}\mathit(f, I_n)} t_{m, c}^n 
	 +  \sum_{{\bf X_c} \in \mathcal{B}_1\mathit(f, I_n)} s_{1, c}^n +  \dots + \sum_{{\bf X_c} \in \mathcal{B}_m\mathit(f, I_n)} s_{m, c}^n \right) 
	 \end{split}
\end{equation*}
\label{lemma:mreductionlowerBound}
\end{lemma}

We also make use of the following lemma which generalizes lemma~\ref{lemma:linearIneq}:

\begin{lemma}
When $d(.)$ is linear, that is, $d(x) = |x|$, the following inequality holds true:
\begin{equation}
\begin{split}
& \frac{1}{L}\sum_{r}\sum_{I_n \in \Gamma_r}\left(\sum_{{\bf X_c} \in \mathcal{A}\mathit(f, I_n)} t_{m, c}^n +  \sum_{{\bf X_c} \in \mathcal{B}_1\mathit(f, I_n)} s_{1, c}^n +  \dots + \sum_{{\bf X_c} \in \mathcal{B}_m\mathit(f, I_n)} s_{m, c}^n \right) \\
& \leq  max \left\{ m \cdot \frac{c}{2}\left(2 + \frac{L}{M}\right), \left(2 + \frac{2M}{L}\right) \right\} \sum_{{\bf c} \in \mathcal{C}} \theta_{\bf c}(\bf {x_c})
\end{split}
\end{equation}
where $\mathit{c}$ is the largest clique in the random field.
\label{lemma:mlinearIneq}
\end{lemma}

Let us denote the set of optimum labels in a clique $\mathbf{X_{c}}$ arranged in an increasing order as ${l_1, l_2....., l_{c-1}, l_c}$. Since we are dealing with the truncated linear metric, the terms $t_{m, c}^n$ and $s_{k, c}^n$ can be simplified as:

\begin{align*}
t_{m, c}^n & = {\bf \omega}_{c}\{(l_c - l_1) + (l_{c - 1} - l_2) + \dots + (l_{c - m + 1} - l_m)\} \\
s_{1, c}^n & = {\bf \omega}_{c}\{(l_c - i_n - 1 + M) + (l_{c - 1} - l_2) + \dots + (l_{c - n + 1}- l_n)\}\\
\vdots \\
s_{m, c}^n & = {\bf \omega}_{c}\{(l_c - i_n - 1 + M) + (l_{c - 1} - i_n - 1 + M) + \dots + (l_{c - n + 1}- i_n - 1 + M)\}\\
\end{align*}

Since the labels are sorted in ascending order $\{l_1, \dots , l_c \}$, it follows that
\begin{align*}
t_{m, c}^n & \leq {\bf \omega}_{c} \cdot m \cdot \{l_c - l_1 \} \\
s_{1, c}^n & \leq {\bf \omega}_{c} \cdot m \cdot \{l_c - i_n - 1 + M\} \\ 
\vdots \\
s_{m, c}^n & \leq {\bf \omega}_{c} \cdot m \cdot \{l_c - i_n - 1 + M\} \\ 
\end{align*}

The LHS of inequality in lemma \ref{lemma:mlinearIneq} can be written as:

\begin{equation}
\begin{split}
& \frac{1}{L}\sum_{{\bf c} \in \mathcal{C}}\left(\sum_{\mathcal{A}\mathit(f, I_n) \ni {\bf c}} t_{m, c}^n + \sum_{\mathcal{B}_1 \mathit(f, I_n) \ni {\bf c}} s_{1, c}^n + \dots + \sum_{\mathcal{B}_m \mathit(f, I_n) \ni {\bf c}} s_{m, c}^n \right) \\
& \leq \frac{1}{L}\sum_{{\bf c} \in \mathcal{C}}\left(\sum_{\mathcal{A}\mathit(f, I_n) \ni {\bf c}} {\bf \omega}_{c} \cdot m \cdot \{l_c - l_1 \} + \sum_{\mathcal{B}_1 \mathit(f, I_n) \ni {\bf c}} {\bf \omega}_{c} \cdot m \cdot \{l_c - i_n - 1 + M\} + \dots \right. \\
&\left. + \sum_{\mathcal{B}_m \mathit(f, I_n) \ni {\bf c}} {\bf \omega}_{c} \cdot m \cdot \{l_c - i_n - 1 + M\}\right) \\
& \leq \frac{1}{L}\sum_{{\bf c} \in \mathcal{C}}\left(\sum_{\mathcal{A}\mathit(f, I_n) \ni {\bf c}} {\bf \omega}_{c} \cdot m \cdot \{l_c - l_1 \} + \sum_{\mathcal{B}_1 \mathit(f, I_n) \cup \dots \cup \mathcal{B}_m \mathit(f, I_n) \ni {\bf c}} {\bf \omega}_{c} \cdot m \cdot \{l_c - i_n - 1 + M\} \right) \\
& = m \cdot \frac{1}{L}\sum_{{\bf c} \in \mathcal{C}}\left(\sum_{\mathcal{A}\mathit(f, I_n) \ni {\bf c}} {\bf \omega}_{c} \cdot t_c^n + \sum_{\mathcal{B} \mathit(f, I_n) \ni {\bf c}} {\bf \omega}_{c} \cdot s_c^n \right) \\
\label{eq:mrewriteIneq}
\end{split}
\end{equation}

The last expression in equation~\ref{eq:mrewriteIneq} is $m$ times expression~\ref{eq:rewriteIneq}. The analysis below follows the same steps as in lemma~\ref{lemma:linearIneq}.

In order to prove the lemma, we consider the following three cases for each clique ${\bf X_c}$: \\

$Case - I: |l_c - l_1| \geq L$ and hence $ \omega_c \cdot M \leq \theta_{\bf c}({\bf{x_c}}) \leq \omega_c \cdot m \cdot M$

This results in 

\begin{align}
m \cdot \frac{1}{L}\sum_{{\bf c} \in \mathcal{C}}\left(\sum_{\mathcal{A}\mathit(f, I_n) \ni {\bf c}} {\bf \omega}_{c} \cdot t_c^n + \sum_{\mathcal{B} \mathit(f, I_n) \ni {\bf c}} {\bf \omega}_{c} \cdot s_c^n \right) &\leq m \cdot \frac{c}{2} \left( 2 + \frac{L}{M} \right) \omega_c \cdot M \nonumber \\
& \leq m \cdot \frac{c}{2} \left( 2 + \frac{L}{M} \right) \theta_{\bf c}({\bf{x_c}})
\label{eq:mcase1bound}
\end{align}

$Case - II: M \leq |l_c - l_1| < L$ and hence $ \omega_c \cdot M \leq \theta_{\bf c}({\bf{x_c}}) \leq \omega_c \cdot m \cdot M$

This gives

\begin{align}
m \cdot \frac{1}{L}\sum_{{\bf c} \in \mathcal{C}}\left(\sum_{\mathcal{A}\mathit(f, I_n) \ni {\bf c}} {\bf \omega}_{c} \cdot t_c^n + \sum_{\mathcal{B} \mathit(f, I_n) \ni {\bf c}} {\bf \omega}_{c} \cdot s_c^n \right) &\leq
m \cdot \left\{\frac{2L}{M} + 2 - \frac{L}{2M} \cdot \left( \frac{c}{c - 1} \right) \right\} \omega_c \cdot M \nonumber \\
& \leq m \cdot \left\{\frac{2L}{M} + 2 - \frac{L}{2M} \cdot \left( \frac{c}{c - 1} \right) \right\}  \theta_{\bf c}({\bf{x_c}})
\label{eq:mcase2bound}
\end{align}

$Case - III: |l_c - l_1| < M$ and hence $\omega_c(l_c - l_1) \leq \theta_{\bf c}({\bf{x_c}}) \leq m \cdot \omega_c(l_c - l_1)$

This leads to 

oo\begin{align}
m \cdot \frac{1}{L}\sum_{{\bf c} \in \mathcal{C}}\left(\sum_{\mathcal{A}\mathit(f, I_n) \ni {\bf c}} {\bf \omega}_{c} \cdot t_c^n + \sum_{\mathcal{B} \mathit(f, I_n) \ni {\bf c}} {\bf \omega}_{c} \cdot s_c^n \right) &\leq m \cdot (2L + 2M)(l_{c} - l_{1}) \nonumber\\
		  & \leq m \cdot  L\left(2 + \frac{2M}{L}\right) \theta_{\bf c}(\bf{x_c})
\label{eq:mcase3bound}
\end{align}

Substituting inequalities \eqref{eq:mcase1bound}, \eqref{eq:mcase2bound} and \eqref{eq:mcase3bound} in expression \eqref{eq:mrewriteIneq}, we obtain inequality of lemma \ref{lemma:mlinearIneq}. This proves the lemma. 
$\blacksquare$

Making use of lemmas \ref{lemma:mreductionlowerBound} and \ref{lemma:mlinearIneq}, the proof of proposition~\ref{prop:mlinearBound} follows exactly the same steps as that of proposition~\ref{prop:linearBound}.

\end{appendices}

\end{document}